# Exploiting Contextual Independence In Probabilistic Inference

**David Poole**  poole@cs.ubc.ca
*Department of Computer Science,*
*University of British Columbia,*
*2366 Main Mall, Vancouver, B.C., Canada V6T 1Z4*
`http://www.cs.ubc.ca/spider/`

**Nevin Lianwen Zhang**  lzhang@cs.ust.hk
*Department of Computer Science,*
*Hong Kong University of Science and Technology, Hong Kong,*
`http://www.cs.ust.hk/~lzhang/`

## Abstract

Bayesian belief networks have grown to prominence because they provide compact representations for many problems for which probabilistic inference is appropriate, and there are algorithms to exploit this compactness. The next step is to allow compact representations of the conditional probabilities of a variable given its parents. In this paper we present such a representation that exploits contextual independence in terms of *parent contexts*; which variables act as parents may depend on the value of other variables. The internal representation is in terms of contextual factors (*confactors*) that is simply a pair of a context and a table. The algorithm, *contextual variable elimination*, is based on the standard variable elimination algorithm that eliminates the non-query variables in turn, but when eliminating a variable, the tables that need to be multiplied can depend on the context. This algorithm reduces to standard variable elimination when there is no contextual independence structure to exploit. We show how this can be much more efficient than variable elimination when there is structure to exploit. We explain why this new method can exploit more structure than previous methods for structured belief network inference and an analogous algorithm that uses trees.

## 1. Introduction

Probabilistic inference is important for many applications in diagnosis, perception, user modelling, and anywhere there is uncertainty about the state of the world from observations. Unfortunately general probabilistic inference is difficult both computationally and in terms of the number of probabilities that need to be specified. Belief (Bayesian) networks (Pearl, 1988) are a representation of independence amongst random variables. They are of interest because the independence is useful in many domains, they allow for compact representations for many practical problems, and there are algorithms to exploit the compact representations. Note that even approximate inference is computationally difficult in the worst case (Dagum and Luby, 1993).

Recently there has been work to extend belief networks by allowing more structured representations of the conditional probability of a variable given its parents (D'Ambrosio, 1995). This has been in terms of either causal independencies (Heckerman and Breese, 1994; Zhang and Poole, 1996), parametric forms such as sigmoidal Bayesian networks (Neal, 1992; Saul, Jaakkola and Jordan, 1996), or by exploiting contextual independencies inherent in stating the conditional probabilities in terms of rules (Poole, 1993) or trees (Smith, Holtzman and Matheson, 1993; Boutilier, Friedman, Goldszmidt and Koller, 1996). In this paper we show how an algorithm that exploits conditional independence for efficient inference in belief networks can be extended to also exploit contextual independence. Poole (1997) provides an earlier, less efficient, version in terms of rules. Zhang and Poole (1999) give an abstract mathematical analysis of how contextual independence can be exploited in inference.

Section 2 introduces belief networks and an algorithm, variable elimination (VE) (Zhang and Poole, 1994) or Bucket Elimination for belief assessment (Dechter, 1996), for computing posterior probabilities in belief that is based on nonlinear dynamic programming (Bertelè and Brioschi, 1972). Section 3 presents a representation for conditional probabilities that lets us state contextual independence in terms of *confactors*. Section 4 shows how the VE algorithm can be extended to exploit the contextual independence in confactors. Section 5 shows how we can improve efficiency by reducing the amount of splitting. Section 6 gives some empirical results on standard and random networks. The details of the experiments are given in Appendix A. Section 7 gives comparisons to other proposals for exploiting contextual independencies. Section 8 presents conclusions and future work.

## 2. Background

In this section we present belief networks and an algorithm, variable elimination, to compute the posterior probability of a set of query variables given some evidence.

### 2.1 Belief Networks

We treat random variables as primitive. We use upper case letters to denote random variables. The domain of a random variable $X$, written $dom(X)$, is a set of values. If $X$ is a random variable and $v \in dom(X)$, we write $X=v$ to mean the proposition that $X$ has value $v$. The function $dom$ can be extended to tuples of variables. We write tuples of variables in upper-case bold font. If $\mathbf{X}$ is a tuple of variables, $\langle X_1, \ldots, X_k \rangle$, then $dom(\mathbf{X})$ is the cross product of the domains of the variables. We write $\langle X_1, \ldots, X_k \rangle = \langle v_1, \ldots, v_k \rangle$ as $X_1 = v_1 \wedge \ldots \wedge X_k = v_k$. This is called an instantiation of $\mathbf{X}$. For this paper we assume there is a finite number of random variables, and that each domain is finite.

We start with a total ordering $X_1, \ldots, X_n$ of the random variables.

**Definition 1** The **parents** of random variable $X_i$, written $\pi_{X_i}$, are a minimal[1] set of the predecessors of $X_i$ in the total ordering such that the other predecessors of $X_i$ are independent of $X_i$ given $\pi_{X_i}$. That is $\pi_{X_i} \subseteq \{X_1, \ldots, X_{i-1}\}$ such that $P(X_i|X_{i-1} \ldots X_1) = P(X_i|\pi_{X_i})$.

A **belief network** (Pearl, 1988) is an acyclic directed graph, where the nodes are random variables[2]. We use the terms node and random variable interchangeably. There is an arc from each element of $\pi_{X_i}$ into $X_i$. Associated with the belief network is a set of probabilities of the form $P(X|\pi_X)$, the conditional probability of each variable given its parents (this includes the prior probabilities of those variables with no parents).

By the chain rule for conjunctions and the independence assumption:

$$P(X_1, \ldots, X_n) = \prod_{i=1}^{n} P(X_i|X_{i-1} \ldots X_1)$$

---

1. If there is more than one minimal set, any minimal set can be chosen to be the parents. There is more than one minimal set only when some of the predecessors are deterministic functions of others.
2. Some people like to say the nodes are *labelled* with random variables. In the definition of a graph, the set of nodes can be any set, in particular, they can be a set of random variables. The set of arcs is a set of ordered pairs of random variables.

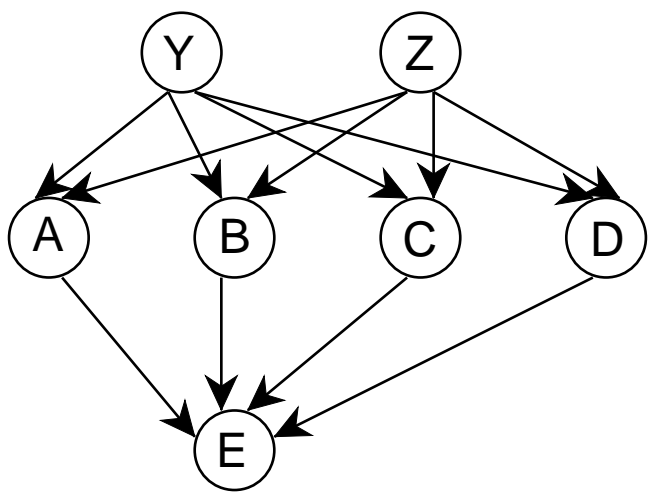

| A | B | C | D | P(e\|ABCD) |
|---|---|---|---|---|
| $a$ | $b$ | $c$ | $d$ | 0.55 |
| $a$ | $b$ | $c$ | $\bar{d}$ | 0.55 |
| $a$ | $b$ | $\bar{c}$ | $d$ | 0.55 |
| $a$ | $b$ | $\bar{c}$ | $\bar{d}$ | 0.55 |
| $a$ | $\bar{b}$ | $c$ | $d$ | 0.3 |
| $a$ | $\bar{b}$ | $c$ | $\bar{d}$ | 0.3 |
| $a$ | $\bar{b}$ | $\bar{c}$ | $d$ | 0.3 |
| $a$ | $\bar{b}$ | $\bar{c}$ | $\bar{d}$ | 0.3 |
| $\bar{a}$ | $b$ | $c$ | $d$ | 0.08 |
| $\bar{a}$ | $b$ | $c$ | $\bar{d}$ | 0.08 |
| $\bar{a}$ | $b$ | $\bar{c}$ | $d$ | 0.025 |
| $\bar{a}$ | $b$ | $\bar{c}$ | $\bar{d}$ | 0.5 |
| $\bar{a}$ | $\bar{b}$ | $c$ | $d$ | 0.08 |
| $\bar{a}$ | $\bar{b}$ | $c$ | $\bar{d}$ | 0.08 |
| $\bar{a}$ | $\bar{b}$ | $\bar{c}$ | $d$ | 0.85 |
| $\bar{a}$ | $\bar{b}$ | $\bar{c}$ | $\bar{d}$ | 0.5 |

Figure 1: A simple belief network and a conditional probability table for $E$.

$$= \prod_{i=1}^{n} P(X_i | \pi_{X_i}) \quad (1)$$

This factorization of the joint probability distribution is often given as the formal definition of a belief network.

**Example 1** Consider the belief network of Figure 1. This represents a factorization of the joint probability distribution:

$P(A, B, C, D, E, Y, Z)$
$= P(E|ABCD)P(A|YZ)P(B|YZ)P(C|YZ)P(D|YZ)P(Y)P(Z)$

If the variables are binary[3], the first term, $P(E|ABCD)$, requires the probability of $E$ for all 16 cases of assignments of values to $A, B, C, D$. One such table is given in Figure 1.

### 2.2 Belief Network Inference

The task of probabilistic inference is to determine the posterior probability of a variable or variables given some observations. In this section we outline a simple algorithm for belief net inference called variable elimination (VE) (Zhang and Poole, 1994; Zhang and Poole, 1996) or bucket elimination for belief assessment (BEBA) (Dechter, 1996), that is based on the ideas of nonlinear dynamic programming (Bertelè and Brioschi, 1972)[4] and is closely related to SPI (Shachter, D'Ambrosio and

---

3. In this and subsequent examples, we assume that variables are Boolean (i.e., with domain {*true, false*}). If $X$ is a variable, $X$=*true* is written as $x$ and $X$=*false* is written as $\bar{x}$, and similarly for other variables. The theory and the implementations are not restricted to binary variables.
4. Bertelè and Brioschi (1972) give essentially the same algorithm, but for the optimization problem of finding a minimization of sums. In VE, we use the algorithm for finding the sum of products. VE is named because of the links to Del Favero, 1990). This is a query oriented algorithm that exploits the conditional independence inherent in the network structure for efficient inference, similar to how clique tree propagation exploits the structure (Lauritzen and Spiegelhalter, 1988; Jensen, Lauritzen and Olesen, 1990).

Suppose we observe variables $E_1, \ldots, E_s$ have corresponding values $o_1 \ldots o_s$. We want to determine the posterior probability of variable $X$, the *query variable*, given evidence $E_1 = o_1 \wedge \ldots \wedge E_s = o_s$:

$$P(X|E_1=o_1 \wedge \ldots \wedge E_s=o_s) = \frac{P(X \wedge E_1=o_1 \wedge \ldots \wedge E_s=o_s)}{P(E_1=o_1 \wedge \ldots \wedge E_s=o_s)}$$

The denominator, $P(E_1=o_1 \wedge \ldots \wedge E_s=o_s)$, is a normalizing factor:

$$P(E_1=o_1 \wedge \ldots \wedge E_s=o_s) = \sum_{v \in dom(X)} P(X=v \wedge E_1=o_1 \wedge \ldots \wedge E_s=o_s)$$

The problem of probabilistic inference can thus be reduced to the problem of computing the probability of conjunctions.

Let $\mathbf{Y} = \{Y_1, \ldots, Y_k\}$ be the non-query, non-observed variables (i.e., $\mathbf{Y} = \{X_1, \ldots, X_n\} - \{X\} - \{E_1, \ldots, E_s\}$). To compute the marginal distribution, we sum out the $Y_i$'s:

$$\begin{aligned} & P(X \wedge E_1=o_1 \wedge \ldots \wedge E_s=o_s) \\ =\ & \sum_{Y_k} \cdots \sum_{Y_1} P(X_1, \ldots, X_n)_{\{E_1=o_1 \wedge \ldots \wedge E_s=o_s\}} \\ =\ & \sum_{Y_k} \cdots \sum_{Y_1} \prod_{i=1}^{n} P(X_i | \pi_{X_i})_{\{E_1=o_1 \wedge \ldots \wedge E_s=o_s\}} \end{aligned}$$

where the subscripted probabilities mean that the associated variables are assigned the corresponding values.

Thus probabilistic inference reduces to the problem of summing out variables from a product of functions. To solve this efficiently we use the distribution law that we learned in high school: to compute a sum of products such as $xy + xz$ efficiently, we distribute out the common factors (which here is $x$) which results in $x(y + z)$. This is the essence of the VE algorithm. We call the elements multiplied together "factors" because of the use of the term in mathematics. Initially the factors represent the conditional probability tables, but the intermediate factors are just functions on variables that are created by adding and multiplying factors.

A **factor** on variables $V_1, \ldots, V_d$ is a representation of a function from $dom(V_1) \times \ldots \times dom(V_d)$ into the real numbers.

Suppose that the $Y_i$'s are ordered according to some elimination ordering. We sum out the variables one at at time.

To sum out a variable $Y_i$ from a product, we distribute all of the factors that don't involve $Y_i$ out of the sum. Suppose $f_1, \ldots, f_k$ are some functions of the variables that are multiplied together (initially these are the conditional probabilities), then

$$\sum_{Y_i} f_1 \ldots f_k = f_1 \ldots f_m \sum_{Y_i} f_{m+1} \ldots f_k$$

---

the algorithm of Bertelè and Brioschi (1972); they refer to their basic algorithm as *The elimination of variables one by one*, which is exactly what we do. Bertelè and Brioschi (1972) also describe good elimination ordering heuristics and refinements such as eliminating variables in blocks and forms of conditioning which we don't consider here.

The only difference between VE and BEBA is that BEBA requires an a priori elimination ordering (and exploits the prior ordering for efficiency), whereas the VE allows for dynamic selection of which variable to eliminate next.

EXPLOITING CONTEXTUAL INDEPENDENCE IN PROBABILISTIC INFERENCE

**To compute** $P(X|E_1=o_1 \wedge \ldots \wedge E_s=o_s)$
    Let $F$ be the factors obtained from the original conditional probabilities.
    1.    Replace each $f \in F$ that involves some $E_i$ with $f_{\{E_1=o_1,\ldots,E_s=o_s\}}$.
    2.    While there is a factor involving a non-query variable
            Select non-query variable $Y$ to eliminate
            Set $F = eliminate(Y, F)$.
    3.    Return $renormalize(F)$

**Procedure** $eliminate(Y, F)$:
    Partition $F$ into
        $\{f_1, \ldots, f_m\}$ that don't contain $Y$ and
        $\{f_{m+1}, \ldots, f_r\}$ that do contain $Y$
    Compute $f = \sum_Y f_{m+1} \otimes_t \ldots \otimes_t f_r$
    Return $\{f_1, \ldots, f_m, f\}$

**Procedure** $renormalize(\{f_1, \ldots, f_r\})$:
    Compute $f = f_1 \otimes_t \ldots \otimes_t f_r$
    Compute $c = \sum_X f$           % $c$ is normalizing constant
    Return $f/c$                 % divide each element of $f$ by $c$

Figure 2: The tabular VE algorithm

where $f_1 \ldots f_m$ are those functions that don't involve $Y_i$, and $f_{m+1} \ldots f_k$ are those that do involve $Y_i$. We explicitly construct a representation for the new function $\sum_{Y_i} f_{m+1} \ldots f_k$, and continue summing out the remaining variables. After all the $Y_i$'s have been summed out, the result is a function on $X$ that is proportional to $X$'s posterior distribution.

In the tabular implementation of the VE algorithm (Figure 2), a function of $d$ discrete variables $V_1, \ldots, V_d$, is represented as a $d$-dimensional table (which can be implemented, for example, as a $d$-dimensional array, as a tree of depth $d$, or, as in our implementation, as a 1-dimensional array based on a lexicographic ordering on the variables). If $f$ is such a table, let $variables(f) = \{V_1, \ldots, V_d\}$. We sometimes write $f$ as $f[V_1, \ldots, V_d]$ to make the variables explicit. $f$ is said to *involve* $V_i$ if $V_i \in variables(f)$.

There are three primitive operations on tables: setting variables, forming the product of tables, and summing a variable from a table.

**Definition 2** Suppose $C$ is a set of variables, $c$ is an assignment $C = v$, and $f$ is a factor on variables $X$. Let $Y = X - C$, let $Z = X \cap C$, and let $Z = v'$ be the assignment of values to $Z$ that assigns the same values to elements of $Z$ as $c$ does. Define $set(f, c)$ be the factor on $Y$ given by:

    $set(f, c)(Y) = f(Y, Z=v')$.

That is, $set(f, c)$ is a function of $Y$, the variables of $f$ that are not in $c$, that is like $f$, but with some values already assigned. Note that, as a special case of this, if $c$ doesn't involve any variable in $f$ then $set(f, c) = f$.

**Example 2** Consider the factor $f(A, B, C, D, E)$ defined by the table of Figure 1. Some examples of the value of this function are $f(a, b, c, d, e) = 0.55$, and $f(\bar{a}, b, c, d, \bar{e}) = 1 - 0.08 = 0.92$. $set(f, \bar{a} \wedge b \wedge e)$ is a function of $C$ and $D$ defined by the table:

| C | D | value |
|---|---|---|
| $c$ | $d$ | 0.08 |
| $c$ | $\bar{d}$ | 0.08 |
| $\bar{c}$ | $d$ | 0.025 |
| $\bar{c}$ | $\bar{d}$ | 0.5 |

**Definition 3** The **product** of tables $f_1$ and $f_2$, written $f_1 \otimes_t f_2$ is a table on the union of the variables in $f_1$ and $f_2$ (i.e., *variables*$(f_1 \otimes_t f_2) =$ *variables*$(f_1) \cup$ *variables*$(f_2)$) defined by:

$$(f_1 \otimes_t f_2)(\mathbf{X}, \mathbf{Y}, \mathbf{Z}) = f_1(\mathbf{X}, \mathbf{Y}) f_2(\mathbf{Y}, \mathbf{Z})$$

where $\mathbf{Y}$ is *variables*$(f_1) \cap$ *variables*$(f_2)$, $\mathbf{X}$ is *variables*$(f_1) -$ *variables*$(f_2)$, and $\mathbf{Z}$ is *variables*$(f_2) -$ *variables*$(f_1)$.

Note that $\otimes_t$ is associative and commutative.

To construct the product of tables, $f_{m+1} \otimes_t \cdots \otimes_t f_k$, we union all of the variables in $f_{m+1} \ldots f_k$, say these are $X_1, \ldots, X_r$. Then we construct an $r$-dimensional table so there is an entry in the table for each combination $v_1, \ldots, v_r$ where $v_i \in dom(X_i)$. The value for the entry corresponding to $v_1, \ldots, v_r$ is obtained by multiplying the values obtained from each $f_i$ applied to the projection of $v_1, \ldots, v_r$ onto the variables of $f_i$.

**Definition 4** The **summing out** of variable $Y$ from table $f$, written $\sum_Y f$ is the table with variables $\mathbf{Z} =$ *variables*$(f) - \{Y\}$ such that[5]

$$(\sum_Y f)(\mathbf{Z}) = \sum_{v_i \in dom(Y)} f(\mathbf{Z} \wedge Y{=}v_i)$$

where $dom(Y) = \{v_1, \ldots, v_s\}$.

Thus, to sum out $Y$, we reduce the dimensionality of the table by one (removing the $Y$ dimension), the values in the resulting table are obtained by adding the values of the table for each value of $Y$.

**Example 3** Consider eliminating $B$ from the factors of Example 1 (representing the belief network of Figure 1), where all of the variables are Boolean. The factors that contain $B$, namely those factors that represent $P(E|ABCD)$ and $P(B|YZ)$, are removed from the set of factors. We construct a factor $f_1(A, B, C, D, E, Y, Z) = P(E|A, B, C, D) \otimes_t P(B|Y, Z)$, thus, for example,

$$f_1(a, b, c, d, e, y, z) = P(e|a \wedge b \wedge c \wedge d)P(b|y \wedge z)$$
$$f_1(\bar{a}, b, c, d, e, y, z) = P(e|\bar{a} \wedge b \wedge c \wedge d)P(b|y \wedge z)$$
$$f_1(\bar{a}, \bar{b}, c, d, e, y, z) = P(e|\bar{a} \wedge \bar{b} \wedge c \wedge d)P(\bar{b}|y \wedge z)$$
$$f_1(a, \bar{b}, c, d, e, y, z) = P(e|a \wedge \bar{b} \wedge c \wedge d)P(\bar{b}|y \wedge z)$$

and similarly for the other values of $A \ldots Z$. We then need to sum out $B$ from $f_1$, producing $f_2(A, C, D, E, Y, Z)$ where, for example,

$$f_2(a, c, d, e, y, z) = f_1(a, b, c, d, e, y, z) + f_1(a, \bar{b}, c, d, e, y, z).$$

$f_2$ is then added to the set of factors. Note that the construction of $f_1$ is for exposition only; we don't necessarily have to construct a table for it explicitly.

---

5. This may look like a circular definition, but the left side defines the summing tables, whereas on the right side we are summing numbers.

## 3. Contextual Independence

In this section we give a formalization of contextual independence. This notion was first introduced into the influence diagram literature (Smith et al., 1993). We base our definitions on the work of Boutilier et al. (1996).

**Definition 5** Given a set of variables $C$, a **context** on $C$ is an assignment of one value to each variable in $C$. Usually $C$ is left implicit, and we simply talk about a context. We would say that $C$ are the variables of the context. Two contexts are **incompatible** if there exists a variable that is assigned different values in the contexts; otherwise they are **compatible**. We write the empty context as *true*.

**Definition 6** (Boutilier et al., 1996) Suppose $X$, $Y$, $Z$ and $C$ are sets of variables. $X$ and $Y$ are **contextually independent** given $Z$ and context $C=c$, where $c \in dom(C)$, if

$$P(X|Y=y_1 \wedge Z=z_1 \wedge C=c) = P(X|Y=y_2 \wedge Z=z_1 \wedge C=c)$$

for all $y_1, y_2 \in dom(Y)$ for all $z_1 \in dom(Z)$ such that $P(Y=y_1 \wedge Z=z_1 \wedge C=c) > 0$ and $P(Y=y_2 \wedge Z=z_1 \wedge C=c) > 0$.

We also say that $X$ is **contextually independent** of $Y$ given $Z$ and context $C=c$. Often we will refer to the simpler case when the set of variables $Z$ is empty; in this case we say that $X$ and $Y$ are **contextually independent** given context $C=c$.

**Example 4** Given the belief network and conditional probability table of Figure 1,

- $E$ is contextually independent of $\{C, D, Y, Z\}$ given context $a \wedge b$.

- $E$ is contextually independent of $\{C, D, Y, Z\}$ given $\{B\}$ and context $a$.

- $E$ is not contextually independent of $\{C, D, Y, Z\}$ given $\{A, B\}$ and the empty context *true*.

- $E$ is contextually independent of $\{B, D, Y, Z\}$ given context $\overline{a} \wedge c$.

- $E$ is contextually independent of $\{A, B, C, D, Y, Z\}$ given $B$ and context $\overline{a} \wedge \overline{c} \wedge d$.

### 3.1 Where Does Contextual Independence Arise?

Most of the examples in this paper are abstract as they are designed to show off the various features of the algorithms or to show pathological cases. In this section we will give some examples to show natural examples. We are not claiming that contextual independence is always present or able to be exploited. Exploiting contextual independence should be seen as one of the tools to solve large probabilistic reasoning tasks.

**Example 5** When a child goes into an emergency ward the staff may want to determine if they are a likely carrier of chicken pox (in order to keep them away from other children). If they haven't been exposed to chicken pox within the previous few weeks, they are unlikely to be a carrier. Thus whether they are a carrier is independent of the other background conditions given they haven't been exposed. If they have been exposed, but have not had chicken pox before they are likely to be a carrier. Thus whether they are a carrier is independent of the other background conditions given they have been exposed and haven't had chicken pox before. The other case can involve many other variables (e.g., the severity and the age of the last time they had chicken pox) to determine how likely the child is to be a carrier.

**Example 6** Many engineered systems are designed to insulate something from other conditions. The classic example is central air conditioning (heating and/or cooling in a house). The temperature inside a house depends on the outside temperature if the air conditioning is off. If the air conditioning is on, the temperature depends on the setting of the thermostat and not on the outside temperature. Thus the inside temperature is contextually independent of the outside temperature given the air conditioning is on and is contextually independent of the thermostat setting given the air conditioning is off.

**Example 7** Consider a case where someone is to make a decision based on a questionnaire and the questions asked depend on previous answers. In this case the decision[6] is contextually independent of the answers to the questions that are not asked given the context of the questions asked. For example, consider a questionnaire to determine if a bank customer should get a loan that starts asking the customer if they rent or own their current home. If they own, they are asked a number of questions about the value of the house which are not asked if they rent. The probability that they get a loan is contextually independent of the value of the home (and the other information that was not available to the decision maker) given that the applicant rents their home.

**Example 8** When learning a decision network from data, it is often advantageous to build a decision tree for each variable given its parents (Friedman and Goldszmidt, 1996; Chickering, Heckerman and Meek, 1997). These decision trees provide contextual independence (a variable is independent of it's predecessors given the context along a path to a leaf in the tree). The reason that this is a good representation to learn is because there are fewer parameters and more fine control over adding parameters; splitting a leaf adds many fewer parameters than adding a new parent (adding a new variable to every context).

### 3.2 Parent Contexts and Contextual Belief Networks

We use the notion of contextual independence for a representation that looks like a belief network, but with finer-grain independence that can be exploited for efficient inference in the contextual variable elimination algorithm.

As in the definition of a belief network, let's assume that we have a total ordering of the variables, $X_1, \ldots, X_n$.

**Definition 7** Given variable $X_i$, we say that $\boldsymbol{C}=\boldsymbol{c}$, where $\boldsymbol{C} \subseteq \{X_{i-1} \ldots X_1\}$ and $c \in dom(\boldsymbol{C})$, is a **parent context** for $X_i$ if $X_i$ is contextually independent of the predecessors of $X_i$ (namely $\{X_{i-1} \ldots X_1\}$) given $\boldsymbol{C}=\boldsymbol{c}$.

What is the relationship to a belief network? In a belief network, the rows of a conditional probability table for a variables form a set of parent contexts for the variable. However, there is often a much smaller set of smaller parent contexts that covers all of the cases.

**Example 9** Consider the belief network and conditional probability table of Figure 1. The predecessors of variable $E$ are $A, B, C, D, Y, Z$. A set of minimal parent contexts for $E$ is $\{\{a,b\}, \{a,\overline{b}\}, \{\overline{a},c\}, \{\overline{a},\overline{c},d,b\}, \{\overline{a},\overline{c},d,\overline{b}\}, \{\overline{a},\overline{c},\overline{d}\}\}$. This is a mutually exclusive and exhaustive set of parent contexts. The probability of $E$ given values for its predecessors can be reduced to the probability of

---

6. To make this a probabilistic problem, and not a decision problem, consider that the probability is for a third party to determine the probability distribution over the possible decisions. A similar analysis can be carried out to exploit contextual independence for decisions (Poole, 1995). The decision maker's decisions can't depend on information she doesn't have.

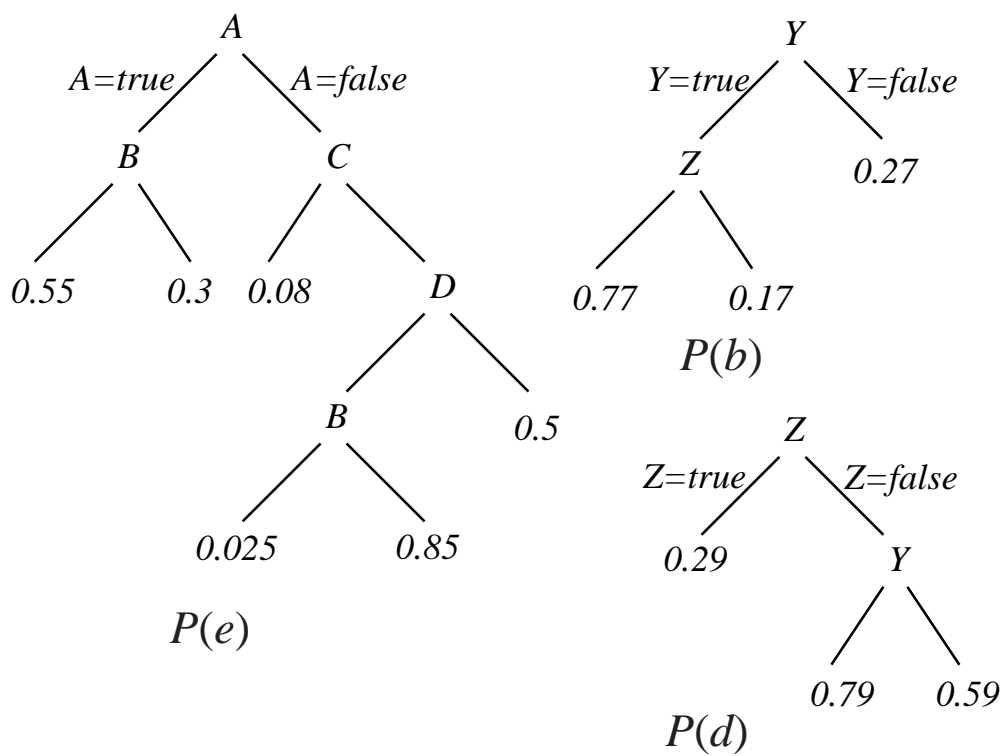

Figure 3: Tree-structured representations of the conditional probabilities for $E$, $B$, and $D$ given their parents. Left branches correspond to true and right branches to false. Thus, for example, $P(e|a \wedge b) = 0.55$, $P(e|a \wedge \bar{b}) = 0.3$, $P(e|\bar{a} \wedge \bar{c} \wedge d \wedge b) = 0.025$ etc.

$E$ given a parent context. For example:

$$P(e|a, b, c, \bar{d}, y, \bar{z}) = P(e|a, b)$$
$$P(e|\bar{a}, b, c, \bar{d}, y, \bar{z}) = P(e|\bar{a}, c)$$
$$P(e|\bar{a}, \bar{b}, c, d, \bar{y}, \bar{z}) = P(e|\bar{a}, c)$$

In the belief network, the parents of $E$ are $A, B, C, D$. To specify the conditional probability of $E$ given its parents, the traditional tabular representation (as in Figure 1) require $2^4 = 16$ numbers instead of the 6 needed if we were to use the parent contexts above. Adding an extra variable as a parent to $E$ doubles the size of the tabular representation, but if it is only relevant in a single context it may only increase the number of parent contexts by one.

We can often (but not always) represent contextual independence in terms of trees. The left side of Figure 3 gives a tree-based representation for the conditional probability of $E$ given its parents. In this tree, internal nodes are labelled with parents of $E$ in the belief network. The left child of a node corresponds to the variable labelling the node being true, and the right child to the variable being false. The leaves are labelled with the probability that $E$ is true. For example $P(e|a \wedge \bar{b}) = 0.3$, irrespectively of the value for $C$ or $D$. In the tree-based representation the variable ($E$ in this case)

POOLE & ZHANGis contextually independent of its predecessors given the context defined by a path through the tree. The paths through the tree correspond to parent contexts.

Before showing how the structure of parent contexts can be exploited in inference, there are a few properties to note:

- The elements of a mutually exclusive and exhaustive set of parent contexts are not always the minimal parent contexts. For example, suppose we have a variable $A$ with parents $B$ and $C$, all of which are Boolean. Suppose probability of $a$ is $p_1$ when both $B$ and $C$ are true and probability $p_2$ otherwise. One mutually exclusive and exhaustive set of parent contexts for $A$ is $\{b \wedge c, b \wedge \overline{c}, \overline{b}\}$. $b \wedge \overline{c}$ is not minimal as $\overline{c}$ is also a parent context. Another mutually exclusive and exhaustive set of parent contexts for this example is $\{b \wedge c, \overline{b} \wedge c, \overline{c}\}$. The set of minimal parent contexts, $\{b \wedge c, \overline{b}, \overline{c}\}$, isn't a mutually exclusive and exhaustive set as the elements are not pairwise incompatible.

  One could imagine using arbitrary Boolean formulae in the contexts. This was not done as it would entail using theorem proving (or a more sophisticated subsumption algorithm) during inference. We doubt that this would be worth the extra overhead for the limited savings.

- A compact decision tree representation of conditional probability tables (Boutilier et al., 1996) always corresponds to a compact set of parent contexts (one context for each path through the tree). However, a mutually exclusive and exhaustive set of parent contexts cannot always be directly represented as a decision tree (as there isn't always a single variable to split on). For example, the mutually exclusive and exhaustive set of contexts $\{\{a, b\}, \{\overline{a}, c\}, \{\overline{b}, \overline{c}\}, \{a, \overline{b}, c\}, \{\overline{a}, b, \overline{c}\}\}$ doesn't directly translate into a decision tree. More importantly, the operations we perform don't necessarily preserve the tree structure. Section 4.12 shows how we can do much better than an analogous tree-based formulation of our inference algorithm.

**Definition 8** A **contextual belief network** is an acyclic directed graph where the nodes are random variables. Associated with each node $X_i$ is a mutually exclusive and exhaustive set of parent contexts, $\Pi_i$, and, for each $\pi \in \Pi_i$, a probability distribution $P(X_i|\pi)$ on $X_i$. Thus a contextual belief network is like a belief network, but we only specify the probabilities for the parent contexts.

For each variable $X_i$ and for each assignment $X_{i-1}{=}v_{i-1}, \ldots, X_1{=}v_1$ of values to its preceding variables, there is a compatible parent context $\pi_{X_i}^{v_{i-1}\ldots v_1}$. The probability of a complete context (an assignment of a value to each variable) is given by:

$$\begin{aligned}
&P(X_1{=}v_1, \ldots, X_n{=}v_n)\\
&= \prod_{i=1}^{n} P(X_i{=}v_n|X_{i-1}{=}v_{i-1}, \ldots, X_1{=}v_1)\\
&= \prod_{i=1}^{n} P(X_i{=}v_i|\pi_{X_i}^{v_{i-1}\ldots v_1})
\end{aligned} \qquad (2)$$

This looks like the definition of a belief network (equation (1)), but which variables act as the parents depends on the values. The numbers required are the probability of each variable for each element of the mutually exclusive and exhaustive set of parent contexts. There can be many fewer of these than the number of assignments to parents in a belief network. At one extreme, there are the same number; at the other extreme there can be exponentially many more assignments of values to parents than the number of elements of a mutually exclusive and exhaustive set of parent contexts.

272

### 3.3 Parent Skeletons

Although the definition of a contextual belief network specifies the contextual independence we want, it doesn't give us a way to organize the parent contexts (in much the same way as a belief network doesn't specify the representation of a conditional probability table). We use the concept of a parent skeleton as a way to organize the parent contexts; we want to use the indexing provided by tables while still allowing for the ability to express context-specific independence.

The notion of a parent context is more fine-grained than that of a parent (the set of parents corresponds to many parent contexts). When there is no context-specific independence, we would like to not have to consider the parent contexts explicitly, but consider just the parents. We will use a parent skeleton to cover both parents and parent contexts as special cases, and to interpolate between them, when the independence depends on some context as well as all values of some other variables.

**Definition 9** A **parent skeletal pair** for variable $X$ is a pair $\langle c, V \rangle$ where $c$ is a context on the predecessors of $X$ and $V$ is a set of predecessors of $X$ such that $X$ is contextually independent of its predecessors given $V$ and context $c$. Note that a parent context is $c \wedge V = v$. A **parent skeleton** for variable $X$ is a set of parent skeletal pairs, $\{\langle c_j, V_j \rangle : 0 < j \leq k\}$, where the $c_j$ are mutually exclusive and exhaustive (i.e., $c_i$ and $c_j$ are incompatible if $i \neq j$, and $\bigwedge_{j=1}^{k} c_j \equiv true$).

**Example 10** A parent skeleton for $E$ from Example 9 is $\{\langle a, \{B\}\rangle, \langle \overline{a} \wedge c, \{\}\rangle, \langle \overline{a} \wedge \overline{c} \wedge d, \{B\}\rangle, \langle \overline{a} \wedge \overline{c} \wedge \overline{d}, \{\}\rangle\}$.

Parent skeletons form the basis of a representation for contextual belief networks. For each variable, $X$, you select a parent skeleton such that for each parent skeleton pair $\langle c_j, V_j \rangle$ in the parent context, $c_j \wedge V_j = v_j$ is a parent context for $X$. For each such parent context pair we specify a probability distribution $P(X | c_j \wedge V_j = v_j)$.

### 3.4 Contextual Factors

Whereas the VE algorithm uses tables both as a representation for conditional probabilities and for the intermediate representations, the contextual variable elimination algorithm defined below uses a hybrid of tables and rules (Poole, 1997) that we call contextual factors or confactors. Confactors cover both tables and rules as special cases.

**Definition 10** A **contextual factor** or **confactor** is a pair of the form:

$$\langle c, t \rangle$$

where $c$ is a context, say $X_1 = v_k \wedge \ldots \wedge X_k = v_k$, and $t$ is a table that represents a function on variables $X_{k+1}, \ldots, X_m$, where $\{X_1, \ldots, X_k\}$ and $\{X_{k+1}, \ldots, X_m\}$ are disjoint sets of variables. $c$ is called the **body** of the confactor and $t$ is called the **table** of the confactor.

A confactor represents a partial function (Zhang and Poole, 1999) from the union of the variables. The function only has a value when the context is true, and the value of the function is obtained by looking up the value in the table.

Just as tables can be used to represent conditional probabilities, confactors can be used to represent conditional probabilities when there is context-specific independence. In particular, a set of parent contexts can be represented as a set of confactors with mutually exclusive and exhaustive bodies. Given a parent skeleton for variable $X$ we can construct a set of confactors for $X$ as follows: for each $\langle c, V \rangle$ in the parent skeleton for $X$, we construct a confactor $\langle c, t(\{X\} \cup V) \rangle$ where $t(\{X = x\} \wedge V = v) = P(X = x | V = v \wedge c)$.

**Definition 11** A confactor is **applicable** on a context if the body of the confactor is compatible with the context.

**Definition 12** Given a confactor $r = \langle X_1 = v_k \wedge \ldots \wedge X_k = v_k, t[X_{k+1}, \ldots, X_m] \rangle$ and a context $c$ that assigns at least the variables $X_1 \ldots X_m$, if $r$ is applicable in $c$, the **value** of the context $c$ with respect to the confactor $r$ is the value of $t[X_{k+1} = v_{k+1}, \ldots, X_m = v_m]$ where $v_{k+1}, \ldots, v_m$ are the values assigned to $X_{k+1}, \ldots, X_m$ in $c$.

**Definition 13** A set $R$ of confactors **represents a conditional probability** $P(X_i | X_1 \ldots X_{i-1})$ if the bodies of the confactors are mutually exclusive and exhaustive, and if $P(X_i = v_i | X_1 = v_1 \wedge \ldots \wedge X_{i-1} = v_{i-1})$ is equal to the value of the context $X_1 = v_1 \wedge \ldots \wedge X_{i-1} = v_{i-1} \wedge X_i = v_i$ with respect to the (unique) confactor in $R$ that is applicable in that context.

Intuitively, the confactors that represent a contextual belief network are a way to organize the parent contexts. The idea is to represent the parent contexts in tables when there is no context-specific independence, and when some variables are independent of their predecessors in some context, then that context can be made a body of the confactors.

**Example 11** Consider the conditional probabilities represented in Figure 3. $E$ is independent of its predecessors given $\{B\}$ and context $a$. This leads to the confactor:

$$\langle a, t_1[B, E] \rangle \tag{3}$$

$E$ is independent of its predecessors given context $\bar{a} \wedge c$. This leads to the confactor:

$$\langle \bar{a} \wedge c, t_2[E] \rangle \tag{4}$$

$E$ is independent of its predecessors given $\{B\}$ and context $\bar{a} \wedge \bar{c} \wedge D$. This leads to the confactor:

$$\langle \bar{a} \wedge \bar{c} \wedge d, t_3[B, E] \rangle \tag{5}$$

$E$ is independent of its predecessors given context $\bar{a} \wedge \bar{c} \wedge \bar{d}$. This leads to the confactor:

$$\langle \bar{a} \wedge \bar{c} \wedge \bar{d}, t_4[E] \rangle \tag{6}$$

The full multiset of confactors corresponding to the trees of Figure 3 are given in Figure 4. The fifth and sixth confactors give the conditional probability for $B$, and the last two confactors give the conditional probability for $D$.

We can now rewrite the definition of a contextual belief network in terms of confactors:

> If every conditional probability is represented by a set of confactors, the probability of a complete context, $c$ is the product of the values of $c$ with respect to the confactors that are applicable in $c$. For each complete context and for each variable there is one confactor containing that variable that is applicable in that context.

## 4. Contextual Variable Elimination

The general idea of contextual variable elimination (CVE) is to represent conditional probabilities in terms of confactors, and use the VE algorithm with the confactor representation rather than with tables. The units of manipulation are thus finer grained than the factors in VE or the members of the buckets of BEBA; what is analogous to a factor or a member of a bucket consists of multisets of confactors. Given a variable to eliminate, we can ignore (distribute out) all of the *confactors* that don't involve this variable. Where there is some contextual independence that goes beyond conditional independence of variables, the savings can be substantial. If there is no contextual independence, all of the confactors have empty contexts, and this algorithm reduces to VE.

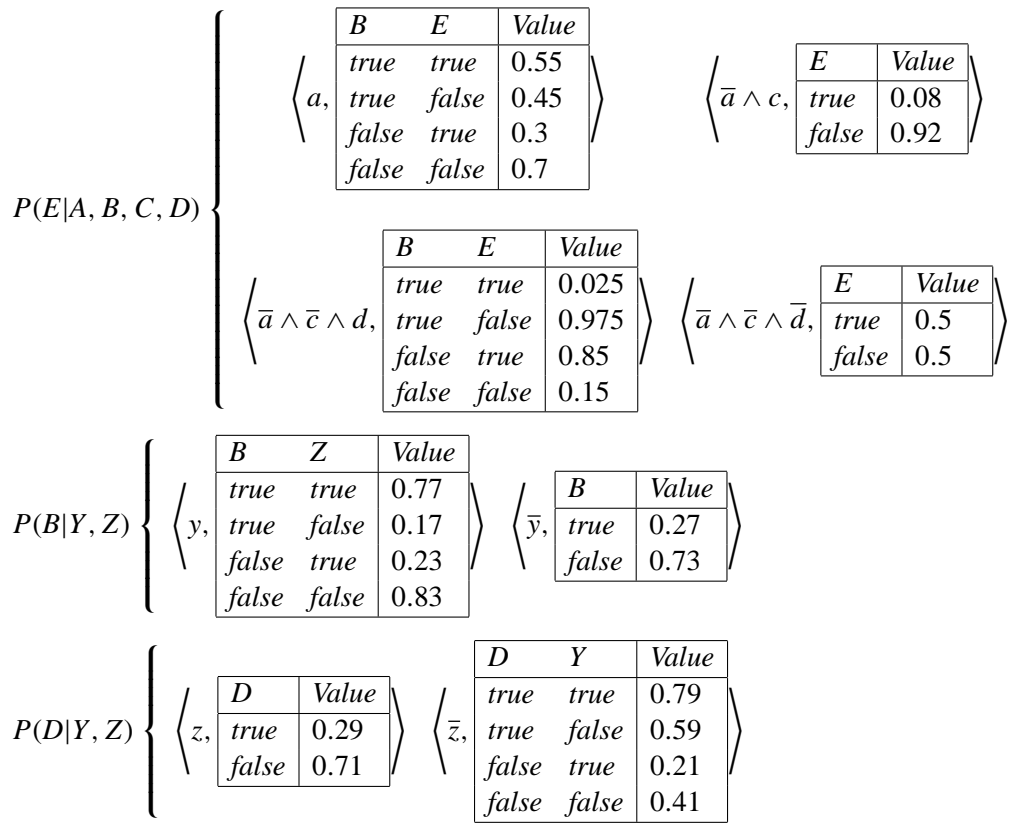

Figure 4: The confactors corresponding to the trees of Figure 3

This section introduces an abstract nondeterministic version of CVE. Section 5 presents a more concrete version where we explain how to resolve much of the nondeterminism.

The input to CVE is:

- a multiset of confactors that consists of the union of the confactors that represent the conditional probability distribution of each variable given its predecessors

- a set of query variables

- an observation that is a conjunction of assignments of values to some of the variables

We first consider the case with no observations. Observations are considered in Section 4.7.

Initially and after the elimination of each variable, we maintain a multiset of confactors with the following **program invariant**:

> The probability of a context $c$ on the non-eliminated variables can be obtained by multiplying the values of context $c$ associated with confactors that are applicable in context $c$. For each complete context on the non-eliminated variables and for each variable there is at least one confactor containing that variable that is applicable in that context[7].

The algorithm will not sum out a variable in all contexts in one step. Rather it will sum out a variable in different contexts separately. Intermediate to being fully summed out, a variable will be summed out in some contexts and not in others. The *remaining variables* should be interpreted relative to whether the variable has been summed out in context $c$.

Like VE, the abstract algorithm is made up of the primitive operations of summing out a variable and multiplying confactors, and also includes a primitive operation of confactor splitting that enables the other two operations. All of these operations locally preserve this program invariant. They are described in the next subsections.

### 4.1 Multiplying Contextual Factors

If we have two confactors with the same context:

$\langle b, t_1 \rangle$

$\langle b, t_2 \rangle$

we can replace them with their product:

$\langle b, t_1 \otimes_t t_2 \rangle$.

---

7. This second part of the invariant may not be so intuitive, but is important. For example, in Example 11, one may be tempted to reduce confactor (6) to $\langle \overline{a} \wedge \overline{c} \wedge \overline{d}, 0.5 \rangle$ (i.e., where the table is a function of no variables) as the contribution of the confactors is the same independent of the value of $E$ (the table $t_4[E]$ has value 0.5 for each value of $E$ in confactor (6)). The first part of the invariant isn't violated. However, if there were no other confactors containing $E$ that are applicable when $\overline{a} \wedge \overline{c} \wedge \overline{d}$ is true, after summing out $E$, we want the confactor $\langle \overline{a} \wedge \overline{c} \wedge \overline{d}, 1 \rangle$, but before summing out $E$ we want the confactor $\langle \overline{a} \wedge \overline{c} \wedge \overline{d}, 0.5 \rangle$ in order to maintain the first part of the invariant. We would like to maintain the property that we only consider confactors containing $E$ when eliminating $E$. The second part of the invariant allows us to do this without treating this as a special case in our algorithm.

The program invariant is maintained, as any context incompatible with $b$ isn't affected by this operation. Any context that is compatible with $b$, the product of the values of $t_1$ and $t_2$ on that context is the same as the value of $t_1 \otimes_t t_2$ on that context. The completeness part of the invariant isn't affected by multiplying.

### 4.2 Summing Out A Variable That Appears In The Table

Suppose we are eliminating $Y$, and have a confactor:

$$\langle b, t \rangle$$

such that table $t$ involves $Y$, and no other confactor that is compatible with $b$ contains $Y$, we can replace this confactor with

$$\langle b, \sum_Y t \rangle$$

Note that after this operation $Y$ is summed out in context $b$.

**Correctness:** To see why this is correct, consider a context $c$ on the remaining variables ($c$ doesn't give a value for $Y$). If $c$ isn't compatible with $b$, it isn't affected by this operation. If it is compatible with $b$, by elementary probability theory:

$$P(c) = \sum_i P(c \wedge Y{=}v_i)$$

By the program invariant, and because there are no other confactors containing $Y$ that are compatible with $c$, $P(c \wedge Y{=}v_i) = p_i p$, for some product $p$ of contributions of confactors that don't involve $Y$. Exactly the same confactors will be used for the different values of $Y$. Thus we have $P(c) = p(\sum_i p_i)$, and so we have maintained the first part of the program invariant. The second part of the program invariant is trivially maintained.

### 4.3 Summing Out A Variable In The Body Of Confactors

Suppose we are eliminating $Y$, with domain $\{v_1, \ldots, v_k\}$, and have confactors:

$$\langle b \wedge Y{=}v_1, T_1 \rangle$$
$$\ldots$$
$$\langle b \wedge Y{=}v_k, T_k \rangle$$

such that there are no other confactors that contain $Y$ whose context is compatible with $b$. We can replace these confactors with the confactor:

$$\langle b, T_1 \oplus_t \ldots \oplus_t T_k \rangle$$

Where $\oplus_t$ is the additive analogue of $\otimes_t$. That is, it follows definition 3, but using addition of the values instead of multiplication.

Note that after this operation $Y$ is summed out in context $b$.

**Correctness:** To see why this is correct, consider a context $c$ on the remaining variables ($c$ doesn't give a value for $Y$). If $c$ isn't compatible with $b$, it isn't affected by this operation. If it is compatible with $b$, by elementary probability theory:

$$P(c) = \sum_i P(c \wedge Y{=}v_i)$$

we can distribute out all of the other confactors from the product and thus the first part of the invariant is maintained. Note that the $\oplus_t$ operation is equivalent to enlarging each table to include the union of all of the variables in the tables, but not changing any of the values, and then pointwise adding the values of the resulting tables. The second part is trivially maintained.

The second part of the program invariant implies that we cannot have a confactor of the form $\langle b \wedge Y{=}v_i, p_i \rangle$ without a corresponding confactor for $Y{=}v_j$, where $i \neq j$.

### 4.4 Confactor Splitting

In order to satisfy the prerequisites to be able to multiply confactors and sum out variables, sometimes we need to split confactors.

If we have a confactor

$\langle b, t \rangle$

we can replace it by the result of splitting it on a non-eliminated variable $Y$, with domain $\{v_1, \ldots, v_k\}$. If $Y$ doesn't appear in $t$, splitting $t$ on $T$ results in the set of confactors:

$\langle b \wedge Y{=}v_1, t \rangle$

$\ldots$

$\langle b \wedge Y{=}v_k, t \rangle$

If $Y$ does appear in $t$, the result is the set of confactors:

$\langle b \wedge Y{=}v_1, set(t, Y{=}v_1) \rangle$

$\ldots$

$\langle b \wedge Y{=}v_k, set(t, Y{=}v_k) \rangle$

where *set* was defined in Definition 2.

**Correctness:** The program invariant is maintained as one of the new confactors is used for any complete context instead of the original confactor. They both give the same contribution.

**Example 12** Splitting the first confactor for $P(E|A, B, C, D)$ in Figure 4 on $Y$ gives two confactors:

$$\left\langle a \wedge y, \begin{array}{|cc|c|} \hline B & E & \text{Value} \\ \hline \text{true} & \text{true} & 0.55 \\ \text{true} & \text{false} & 0.45 \\ \text{false} & \text{true} & 0.3 \\ \text{false} & \text{false} & 0.7 \\ \hline \end{array} \right\rangle \quad (7)$$

$$\left\langle a \wedge \bar{y}, \begin{array}{|cc|c|} \hline B & E & \text{Value} \\ \hline \text{true} & \text{true} & 0.55 \\ \text{true} & \text{false} & 0.45 \\ \text{false} & \text{true} & 0.3 \\ \text{false} & \text{false} & 0.7 \\ \hline \end{array} \right\rangle \quad (8)$$

**Example 13** Splitting the first confactor for $P(B|Y, Z)$ in Figure 4 on $A$ gives two confactors:

$$\left\langle a \wedge y, \begin{array}{|cc|c|} \hline B & Z & Value \\ \hline true & true & 0.77 \\ true & false & 0.17 \\ false & true & 0.23 \\ false & false & 0.83 \\ \hline \end{array} \right\rangle \quad (9)$$

$$\left\langle \bar{a} \wedge y, \begin{array}{|cc|c|} \hline B & Z & Value \\ \hline true & true & 0.77 \\ true & false & 0.17 \\ false & true & 0.23 \\ false & false & 0.83 \\ \hline \end{array} \right\rangle \quad (10)$$

The reason that we may want to do these two splits is that now we can multiply confactors (7) and (9).

### 4.5 Examples of Eliminating Variables

The four operations above are all that is needed to eliminate a variable. A variable is eliminated when it is summed out of all contexts.

**Example 14** When we eliminate $B$ from the confactors of Figure 4, we only need to consider the four confactors that contain $B$. The preconditions for summing out $B$ or for multiplying are not satisfied, so we need to split. If we split the first confactor for $P(E|A, B, C, D)$ on $Y$ (as in Example 12) and split the first confactor for $P(B|Y, Z)$ on $A$ (as in Example 13), we produce two confactors, (7) and (9), that can be multiplied producing:

$$\left\langle a \wedge y, \begin{array}{|ccc|c|} \hline B & E & Z & Value \\ \hline true & true & true & 0.4235 \\ true & true & false & 0.0935 \\ true & false & true & 0.3465 \\ true & false & false & 0.0765 \\ false & true & true & 0.069 \\ false & true & false & 0.249 \\ false & false & true & 0.161 \\ false & false & false & 0.581 \\ \hline \end{array} \right\rangle \quad (11)$$

This is the only confactor that contains $B$ and is applicable in the context $a \wedge y$, so we can sum out $B$ from the table, producing the confactor:

$$\left\langle a \wedge y, \begin{array}{|cc|c|} \hline E & Z & Value \\ \hline true & true & 0.4925 \\ true & false & 0.3425 \\ false & true & 0.5075 \\ false & false & 0.6575 \\ \hline \end{array} \right\rangle \quad (12)$$

The other nontrivial confactors produced when summing out $B$ are:

$$\left\langle a \wedge \bar{y}, \begin{array}{|c|c|} \hline E & Value \\ \hline true & 0.3675 \\ false & 0.6325 \\ \hline \end{array} \right\rangle \quad (13)$$

$$\left\langle \bar{a} \wedge \bar{c} \wedge d \wedge y, \begin{array}{|cc|c|} \hline E & Z & \text{Value} \\ \hline \text{true} & \text{true} & 0.21475 \\ \text{true} & \text{false} & 0.70975 \\ \text{false} & \text{true} & 0.78525 \\ \text{false} & \text{false} & 0.29025 \\ \hline \end{array} \right\rangle \tag{14}$$

$$\left\langle \bar{a} \wedge \bar{c} \wedge d \wedge \bar{y}, \begin{array}{|c|c|} \hline E & \text{Value} \\ \hline \text{true} & 0.62725 \\ \text{false} & 0.37275 \\ \hline \end{array} \right\rangle \tag{15}$$

See Example 19 below for some trivial confactors produced and how to avoid them.

These confactors should be contrasted with the factor on $A, C, D, E, Y, Z$ (of size 32) that is produced by eliminating $B$ in VE.

**Example 15** Suppose that instead we were to eliminate $D$ from the confactors of Figure 4. This example differs from the previous example as $D$ appear in the bodies as well as in the tables.

The two confactors for $P(E|A, B, C, D)$ that contain $D$, namely $\langle \bar{a} \wedge \bar{c} \wedge d, t_3[B, E] \rangle$ (confactor (5)), and $\langle \bar{a} \wedge \bar{c} \wedge \bar{d}, t_4[E] \rangle$ (confactor (6)) are both compatible with both confactors for $P(D|Y, Z)$. So we cannot sum out the variable or multiply any confactors.

In order to be able to multiply confactors, we can split confactor (5) on $Z$ producing:

$$\langle \bar{a} \wedge \bar{c} \wedge d \wedge z, t_3[B, E] \rangle \tag{16}$$

$$\langle \bar{a} \wedge \bar{c} \wedge d \wedge \bar{z}, t_3[B, E] \rangle \tag{17}$$

The confactors for $P(D|Y, Z)$ are $\langle z, t_7[D] \rangle$ and $\langle z, t_8[D, Y] \rangle$. We can split the first of these on $A$ producing

$$\langle a \wedge z, t_7[D] \rangle \tag{18}$$

$$\langle \bar{a} \wedge z, t_7[D] \rangle \tag{19}$$

There are no other confactors containing $D$ with context compatible with confactor (18). The prerequisite required to sum out $D$ in the context $a \wedge z$ is satisfied. This results in the confactor $\langle a \wedge z, 1 \rangle$ where 1 is the factor of no variables that has value 1. This can be removed as the product of 1 doesn't change anything. Intuitively this can be justified because in the context when $A$ is true $D$ has no children. We can detect this case to improve efficiency (see Section 4.10).

The confactor (19) can be split on $C$, producing

$$\langle \bar{a} \wedge c \wedge z, t_7[D] \rangle \tag{20}$$

$$\langle \bar{a} \wedge \bar{c} \wedge z, t_7[D] \rangle \tag{21}$$

We can sum out $D$ from confactor (20), producing $\langle \bar{a} \wedge c \wedge z, 1 \rangle$, as in the previous case.

We can split confactor (21) on $D$ producing:

$$\langle \bar{a} \wedge \bar{c} \wedge d \wedge z, 0.29 \rangle \tag{22}$$

$$\langle \bar{a} \wedge \bar{c} \wedge \bar{d} \wedge z, 0.71 \rangle \tag{23}$$

where 0.29 and 0.71 are the corresponding values from $t_7[D]$. These are functions of no variables, and so are just numbers.

We can now multiply confactor (22) and (16), producing:

$$\langle \bar{a} \wedge \bar{c} \wedge d \wedge z, 0.29 t_3[B, E] \rangle \tag{24}$$

where $0.29 t_3[B, E]$ is the table obtained by multiplying each element of $t_3[B, E]$ by 0.29.

We can also split confactor (6) on $Z$, producing:

$$\langle \bar{a} \wedge \bar{c} \wedge \bar{d} \wedge z, t_4[E] \rangle \tag{25}$$

$$\langle \bar{a} \wedge \bar{c} \wedge \bar{d} \wedge \bar{z}, t_4[E] \rangle \tag{26}$$

We can multiply confactors (23) and (25), producing:

$$\langle \bar{a} \wedge \bar{c} \wedge \bar{d} \wedge z, 0.71 t_4[E] \rangle \tag{27}$$

We now have only complementary confactors for $D$ in the context $\bar{a} \wedge \bar{c} \wedge z$, namely confactors (24) and (27) so we can sum-out $D$ in this context resulting in

$$\langle \bar{a} \wedge \bar{c} \wedge z, t_9[B, E] \rangle \tag{28}$$

where $t_9[B, E]$ is $0.29 t_3[B, E] \oplus_t 0.71 t_4[E]$. In full form this is:

$$\left\langle \bar{a} \wedge \bar{c} \wedge z, \begin{array}{|cc|c|} \hline B & E & \text{Value} \\ \hline \text{true} & \text{true} & 0.36225 \\ \text{true} & \text{false} & 0.63775 \\ \text{false} & \text{true} & 0.6015 \\ \text{false} & \text{false} & 0.3985 \\ \hline \end{array} \right\rangle \tag{29}$$

The other confactor produced when summing out $D$ is:

$$\left\langle \bar{a} \wedge \bar{c} \wedge \bar{z}, \begin{array}{|ccc|c|} \hline B & E & Y & \text{Value} \\ \hline \text{true} & \text{true} & \text{true} & 0.12475 \\ \text{true} & \text{true} & \text{false} & 0.21975 \\ \text{true} & \text{false} & \text{true} & 0.87525 \\ \text{true} & \text{false} & \text{false} & 0.78025 \\ \text{false} & \text{true} & \text{true} & 0.7765 \\ \text{false} & \text{true} & \text{false} & 0.7065 \\ \text{false} & \text{false} & \text{true} & 0.2235 \\ \text{false} & \text{false} & \text{false} & 0.2935 \\ \hline \end{array} \right\rangle \tag{30}$$

### 4.6 When to Split

Confactor splitting makes the multiset of confactors more complicated, so we have to be careful to apply this operation judiciously. We need to carry out confactor splitting in order to make identical or complementary contexts so we can carry out the operations of summing out a variable or multiplying confactors. These are the only cases we need to split.

**Definition 14** Given confactor $r_1 = \langle c_1, T_1 \rangle$ and context $c$, such that $c_1$ and $c$ are compatible, to **split $r_1$ on $c$** means to split $r_1$ sequentially on each of the variables that are assigned in $c$ that aren't assigned in $c_1$.

When we split $r_1$ on $c$, we end up with a single confactor with a context that is compatible with $c$; the contexts of all of the other confactors that are produced by the splitting are incompatible with $c$. These confactors that are incompatible with $c$ are called **residual** confactors.

More formally, we can recursively define $residual(r_1, c)$, where $r_1 = \langle c_1, t_1 \rangle$ and $c$ and $c_1$ are compatible, by:

- $residual(r_1, c) = \{\}$ if $c \subseteq c_1$

- Else if $c \not\subseteq c_1$, select a variable $X$ that is assigned in $c$ but not in $c_1$.

$$residual(r_1, c) = \{\langle c_1 \wedge X{=}v_i, set(t_1, X{=}v_i)\rangle : v_i \in dom(X) \& v_i \neq c^X\}$$
$$\cup residual(\langle c_1 \wedge X{=}c^X, set(t_1, X{=}c^X)\rangle, c)$$

where $c^X$ is the value assigned to $X$ in context $c$. Recall (Definition 2) that $set(t, X{=}v_i)$ is $t$ if $t$ doesn't involve $X$ and is the selection of the $X{=}v_i$ values from the table, followed by the projection onto the remaining variables, if $t$ does involve $X$.

The results of splitting a confactor on a context is a set of confactors:

$$split(\langle c_1, t_1 \rangle, c) = residual(\langle c_1, t_1 \rangle, c) \cup \{\langle c_1 \cup c, t_1 \rangle\}.$$

**Example 16** Consider $residual(\langle a \wedge b, t_1[C, D]\rangle, c \wedge e)$. Suppose we split on $C$ first, then on $E$. This results in two residual confactors: $\langle a \wedge b \wedge \overline{c}, t_2[D]\rangle$ and $\langle a \wedge b \wedge c \wedge \overline{e}, t_3[D]\rangle$. Note that $t_2[D]$ is the projection of $t_1[C, D]$ onto $C{=}false$ and $t_3[D]$ is the projection of $t_1[C, D]$ onto $C{=}true$. The non-residual confactor that we want from the split is $\langle a \wedge b \wedge c \wedge e, t_3[D]\rangle$.

If instead we split on $E$ then $C$, we get the residual confactors: $\langle a \wedge b \wedge \overline{e}, t_1[C, D]\rangle$ and $\langle a \wedge b \wedge \overline{c} \wedge e, t_2[D]\rangle$, with the same non-residual confactor.

Note that the result can depend on the order in which variables are selected (see below for some useful splitting heuristics). The algorithms that use the split will be correct no matter which order the variables are selected, however some orderings may result in more splitting in subsequent operations.

Example 16 highlights one heuristic that seems generally applicable. When we have to split a confactor on variables that appear in its body and on variables in its table, it's better to split on variables in the table first, as these simplify the confactors that need to be subsequently split.

We can use the notion of a residual to split two rules that are compatible, and need to be multiplied. Suppose we have confactors $r_1 = \langle c_1, t_1 \rangle$ and $r_2 = \langle c_2, t_2 \rangle$, that both contain the variable being eliminated and where $c_1$ and $c_2$ are compatible contexts. If we split $r_1$ on $c_2$, and split $r_2$ on $c_1$, we end up with two confactors whose contexts are identical. Thus we have the prerequisite needed for multiplying.

**Example 17** Suppose we have confactors $r_1 = \langle a \wedge b \wedge \overline{c}, t_1 \rangle$ and $r_2 = \langle a \wedge d, t_2 \rangle$ that both contain the variable being eliminated. We can split $r_1$ on the body of $r_2$, namely $a \wedge d$, producing the confactors

$$\langle a \wedge b \wedge \overline{c} \wedge d, t_1 \rangle \tag{31}$$
$$\langle a \wedge b \wedge \overline{c} \wedge \overline{d}, t_1 \rangle$$

Only the first of these is compatible with $r_2$. The second confactor is a residual confactor.

We can split $r_2$ on the body of $r_1$, namely $a \wedge b \wedge \overline{c}$, by first splitting $r_2$ on $B$, then on $C$, producing the confactors:

$$\langle a \wedge b \wedge c \wedge d, t_2 \rangle$$
$$\langle a \wedge b \wedge \overline{c} \wedge d, t_2 \rangle \tag{32}$$
$$\langle a \wedge \overline{b} \wedge d, t_2 \rangle$$

Only the second confactor (confactor (32)) is compatible with $r_1$ or any of the residual confactors produced by splitting $r_1$. Confactors (31) and (32) have identical contexts and so can be multiplied.

Suppose we have confactors $r_1 = \langle c_1 \wedge Y{=}v_i, t_1 \rangle$ and $r_2 = \langle c_2 \wedge Y{=}v_j, t_2 \rangle$, where $c_1$ and $c_2$ are compatible contexts, and $v_i \neq v_j$. If we split $r_1$ on $c_2$, and split $r_2$ on $c_1$, we end up with two confactors whose contexts are identical except for the complementary values for $Y$. This is exactly what we need for summing out $Y$.

If $Y$ is binary with domain $\{v_i, v_j\}$, and there are confactors $r_1 = \langle c_1 \wedge Y{=}v_i, t_1 \rangle$ and $r_2 = \langle c_2 \wedge Y{=}v_j, t_2 \rangle$, where $c_1$ and $c_2$ are compatible contexts, and there is no other confactor that contains $Y$ that is compatible with $c_1$ and $c_2$, summing out $Y$ in the context $c_1 \cup c_2$ results in the confactors:

$$residual(r_1, c_2) \cup residual(r_2, c_1) \cup \{\langle c_1 \cup c_2, t_1 \oplus_t t_2 \rangle\}.$$

If there are more than two values in the domain, we may need to split each pair of confactors, always using the results of previous splits for subsequent splits.

**Proposition 1** Splitting confactor $\langle c_1, t_1 \rangle$ on $c$ creates

$$\sum_{X \in vars(c) - vars(c_1)} (|dom(X)| - 1)$$

extra confactors, independently of the order in which the variables are selected to be split, where $vars(c)$ is the set of variables assigned in context $c$.

When we have to split, there is a choice as to which variable to split on first. While this choice does not influence the number of confactors created for the single split, it can influence the number of confactors created in total because of subsequent splitting. One heuristic was given above. Another useful heuristic seems to be: given a confactor with multiple possible splits, look at all of the confactors that need to be combined with this confactor to enable multiplication or addition, and split on the variable that appears most. For those cases where the conditional probability forms a tree structure, this will tend to split on the root of the tree first.

### 4.7 Evidence

As in VE, evidence simplifies the knowledge base. Suppose $E_1{=}o_1 \wedge \ldots \wedge E_s{=}o_s$ is observed. There are three steps in absorbing evidence:

- Remove any confactor whose context contains $E_i{=}o'_i$, where $o_i \neq o'_i$.

- Remove any term $E_i{=}o_i$ in the context of a confactor.

- Replace each table $t$ with $set(t, E_1{=}o_1 \wedge \ldots \wedge E_s{=}o_s)$ (as in the tabular VE algorithm).

Again note that incorporating evidence only simplifies the confactor base.

Once evidence has been incorporated into the confactor-base, the **program invariant** becomes:

> The probability of the evidence conjoined with a context $c$ on the non-eliminated, non-observed variables is equal to the product of the probabilities of the confactors that are applicable in context $c$. For each context $c$ on the non-eliminated, non-observed variables and for each variable $X$ there is at least one confactor containing $X$ that is applicable in context $c$.

For probabilistic inference, where we will normalise at the end, we can remove any confactor that doesn't involve any variable (i.e., with an empty context and single number as the table) as a result of the second or third cases. That is, we remove any confactor that only has observed variables. We then need to replace "equal" with "proportional" in the program invariant.

**Example 18** Suppose $\bar{d} \wedge \bar{z}$ is observed given the confactors of Figure 4. The first two confactors for $P(E|A, B, C, D)$ don't involve $D$ or $Z$ and so are not affected by the observation. The third confactor is removed as its body is incompatible with the observation. The fourth confactor is replaced by:

$$\left\langle \bar{a} \wedge \bar{c}, \begin{array}{|c|c|} \hline E & Value \\ \hline true & 0.5 \\ \hline false & 0.5 \\ \hline \end{array} \right\rangle$$

The first confactor for $P(B|Y, Z)$ is replaced by

$$\left\langle y, \begin{array}{|c|c|} \hline B & Value \\ \hline true & 0.17 \\ \hline false & 0.83 \\ \hline \end{array} \right\rangle$$

The first confactor for $P(D|Y, Z)$ is removed and the second is replaced by

$$\left\langle true, \begin{array}{|c|c|} \hline Y & Value \\ \hline true & 0.21 \\ \hline false & 0.41 \\ \hline \end{array} \right\rangle$$

where *true* represents the empty context.

### 4.8 Extracting the Answer

Suppose we had a single query variable $X$. After setting the evidence variables, and eliminating the remaining variables, we end up with confactors of the form:

$$\langle \{X = v_i\}, p_i \rangle$$

and of the form

$$\langle \{\}, t_i[X] \rangle$$

If $e$ is the evidence the probability of $X = v_i \wedge e$ is proportional to the product contributions of the confactors with context $X = v_i$ and the selection for the $X = v_i$ value for the table. Thus

$$P(X = v_i \wedge e) \propto \prod_{\langle X = v_i, p_i \rangle} p_i \prod_{\langle \{\}, t_i[X] \rangle} t_i[v_i].$$

Then we have:

$$P(X = v_i | e) = \frac{P(X = v_i \wedge e)}{\sum_{v_j} P(X = v_j \wedge e)}.$$

Notice that constants of proportionality of the evidence or by removing constants (confactors with no variables) cancel in the division.

If we had multiple query variables (i.e., we wanted the marginal of the posterior), then we still multiply the remaining confactors and renormalise.

### 4.9 The Abstract Contextual Variable Elimination Algorithm

The contextual variable elimination algorithm, is given in Figure 5. A more refined version that does less splitting is given in Section 5.

The elimination procedure is called once for each non-query, non-observed variable. The order in which the variables are selected is called the **elimination ordering**. This algorithm does not imply

**To compute** $P(X|E_1{=}o_1 \wedge \ldots \wedge E_s{=}o_s)$
    **Given** multiset $R$ of confactors
  1.    Incorporate evidence as in Section 4.7.
  2.    **while** there is a non-query variable to be eliminated
        {
            Select non-query variable $Y$ to eliminate;
            Call $R := \textit{eliminate}(Y, R)$;
        }
  3.    Compute posterior probability for $X$ as in Section 4.8

**Procedure** *eliminate*$(Y, R)$:
    **partition** $R$ into:
        $R^-$ = those confactors in $R$ that don't involve $Y$;
        $R^* = \{r \in R : r \text{ involves } Y\}$;
    **while** there is $\{\langle b_1, T_1\rangle, \langle b_2, T_2\rangle\} \subseteq R^*$ where $b_1$ and $b_2$ are compatible,
        {
            remove $\langle b_1, T_1\rangle$ and $\langle b_2, T_2\rangle$ from $R^*$;
            split $\langle b_1, T_1\rangle$ on $b_2$ putting residual confactors in $R^*$;
            split $\langle b_2, T_2\rangle$ on $b_1$, putting residual confactors in $R^*$;
            add $\langle b_1 \wedge b_2, T_1 \otimes_t T_2\rangle$ to $R^*$;
        }
    **for every** $\langle b, t\rangle \in R^*$ such that $Y$ appears in $t$
        {
            remove $\langle b, t\rangle$ from $R^*$;
            add $\langle b, \sum_Y t\rangle$ to $R^-$;
        }
    **while** $R^*$ is not empty
        {
            **if** $\{\langle b \wedge Y{=}v_1, T_1\rangle, \ldots, \langle b \wedge Y{=}v_k, T_k\rangle\} \subseteq R^*$
                {
                    remove $\langle b \wedge Y{=}v_1, T_1\rangle, \ldots, \langle b \wedge Y{=}v_k, T_k\rangle$ from $R^*$;
                    add $\langle b, T_1 \oplus_t \ldots \oplus_t T_k\rangle$ to $R^-$;
                }
            **else if** $\{\langle b_1 \wedge Y{=}v_i, T_1\rangle, \langle b_2 \wedge Y{=}v_j, T_2\rangle\} \subseteq R^*$ where $b_1$ and $b_2$ are compatible and $b_1 \neq b_2$
                {
                    remove $\langle b_1 \wedge Y{=}v_i, T_1\rangle$ and $\langle b_2 \wedge Y{=}v_j, T_2\rangle$ from $R^*$;
                    split $\langle b_1 \wedge Y{=}v_i, T_1\rangle$ on $b_2$, putting all created confactors in $R^*$;
                    split $\langle b_2 \wedge Y{=}v_j, T_2\rangle$ on $b_1$, putting all created confactors in $R^*$;
                }
        }
    **Return** $R^-$.

$\otimes_t$ is defined in Section 2.2.
$\oplus_t$ is defined in Section 4.3.
All set operations are assumed to be on multisets.

Figure 5: Contextual Variable Elimination

that the elimination ordering has to be given a priori. The other choice points are the order in which to do multiplication, and the splitting ordering.

Note that in the *eliminate* algorithm, all set operations are assumed to be on multisets. It is possible, and not uncommon, to get multiple copies of the same confactor. One example where this happens is when there is a naive Bayes model with variable $C$ with no parents, and variables $Y_1, \ldots, Y_n$ each with only $C$ as a parent. Often the conditional probabilities of some of the $Y_i$'s are the same as they represent repeated identical sensors. If these identical sensors observe the same value, then we will get identical confactors, none of which can be removed without affecting the answer.

To see the correctness of the procedure, note that all of the local operations preserve the program invariants; we still need to check that the algorithm halts. After the first while-loop of eliminate, the contexts of the confactors in $R^*$ are mutually exclusive and covering by the second part of the loop invariant. For all complete contexts on the variables that remain after $Y$ is eliminated, there is either a compatible confactor with $Y$ in the table, or there is a compatible confactor with $Y = v_i$ for every value $v_i$. The splitting of the second while loop of eliminate preserves the mutual exclusiveness of the bodies of the confactors in $R^*$ and when splitting a confactor, the set of created confactors covers the same context as the original confactor. If there are confactors in $R^*$, and the if-condition does not hold, then there must be a pair of confactors where the else-if condition holds. Thus, each time through the second while-loop, the number of confactors in $R^-$ increases or the number of confactors in $R^*$ increases and these are both bounded in size by the size of the corresponding factor. Thus *eliminate* must stop, and when it does $Y$ is eliminated in all contexts.

### 4.10 Ones

In a Bayesian network we can remove a non-observed, non-query node with no children without changing the conditional probability of the query variable. This can be carried out recursively. In VE, if we were to eliminate such variables, we create factors that are all ones (as $\sum_X P(X|Y) = 1$).

In contextual VE, we can have a more subtle version when a variable may have no children in some contexts, even if it has children in another context.

**Example 19** Consider eliminating $B$ as in Example 14 where the belief network is given in Figure 1 and the structured conditional probabilities are given in Figure 3). In the context $\bar{a} \wedge c$, the only confactors that are applicable are those that define $P(B|YZ)$. As stated, the contextual VE algorithm, the following three confactors are created:

$\langle \bar{a} \wedge c \wedge y, 1[Z] \rangle$
$\langle \bar{a} \wedge c \wedge \bar{y}, 1 \rangle$

where $1[Z]$ is a function of $Z$ that has value 1 everywhere and 1 is the function of no variables that has value 1.

Confactors that have contribution 1 can be removed without affecting the correctness of the algorithms (as long as these confactors aren't the only confactors that contain a variable in some context). It is easy to show that the first part of the program invariant is maintained as multiplying by 1 doesn't affect any number. The second part of the invariant is also maintained, as there are always the confactors for the child ($E$ in this case) that don't depend on the variable being eliminated, as well as the confactors for the parents of the variable being eliminated.

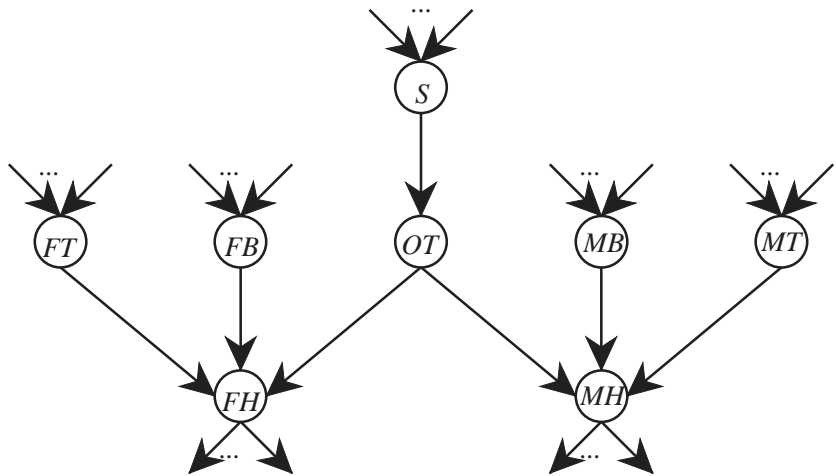

Figure 6: A fragment of a belief network: *OT* to eliminate.

It is however probably better to never construct such confactors rather than to construct them and then throw them away. We show how this can be done in Section 5.1.

### 4.11 Multi-Valued Variables

We have presented an algorithm that allows for multi-valued variables, where the splitting operation creates the same number of confactors as there are values in the domain of the variable being split.

There is an alternate method for using multi-valued variables. This is to extend the notion of a context to allow for membership in a set of values. That is, a context could be a conjunction of terms of the form $X \in S$ where $S$ is a set of values. The original version of equality of the form $X=v$ is the same as $X \in \{v\}$. Splitting can be done more efficiently as there is only one residual confactor for each split. Effectively we treat a multiset of confactors as a unit.

There are examples where this representation can be much more efficient, but it makes the algorithm much more complicated to explain. There are also examples where the binary splitting is less efficient as it needs more splits to get the same result.

### 4.12 Why CVE Does More Than Representing Factors As Trees

It may seem that CVE is a confactor based representation for factors, in much the same way as the trees in the structured policy iteration (Boutilier, Dearden and Goldszmidt, 1995) for solving MDPs. In this section we present a detailed example that explains why CVE can be much more efficient than a tree-based representation of the VE factors.

**Example 20** Figure 6 shows a fragment of a belief network. This is an elaboration of Example 6, where what affects the inside temperature depends on whether the air conditioning is broken or is working. All the variables are Boolean. We use the following interpretation:

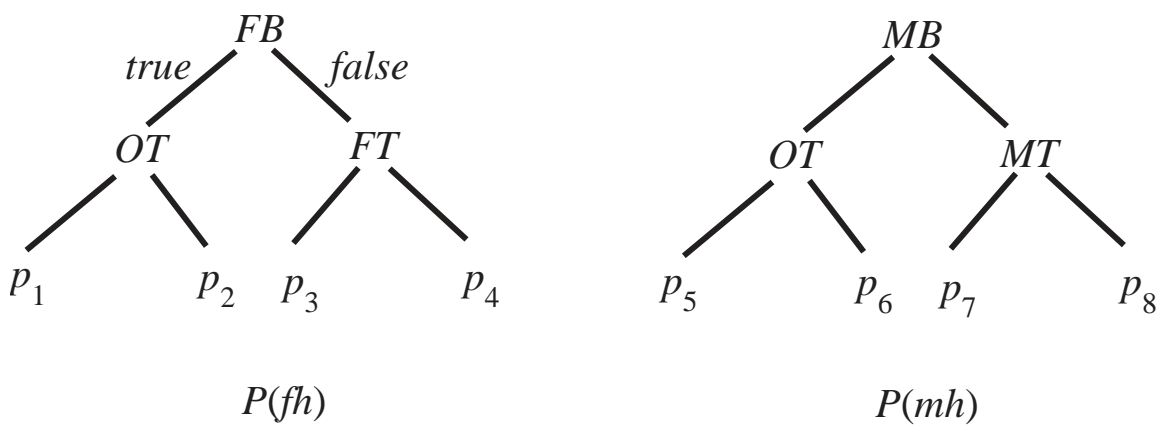

Figure 7: Tree-structured conditional probability tables for *A* and for *B*. Left branches correspond to *true* and right branches to *false*. Thus $p_1 = P(a|d \wedge e)$, $p_2 = P(a|d \wedge \bar{e})$, etc.

*FB* Fred's air conditioning is broken

*FT* Fred's thermostat setting is high

*OT* Outside temperature is hot

*FH* Fred's house is hot

*MB* Mary's air conditioning is broken

*MT* Mary's thermostat setting is high

*MH* Mary's house is hot

*S* Season, that is true if it is Summer

The ancestors of *FT*, *FB*, *S*, *MB*, and *MT* are not shown. They can be multiply connected. Similarly, the descendants of *FH* and *MH* are not shown. They can be multiply connected.

The outside temperature (*OT*) is only relevant to Fred's house being hot (*FH*) when Fred's air conditioner is broken (*FB* is true) in which case Fred's thermostat setting (*FT*) is not relevant. Fred's thermostat setting (*FT*) is only relevant to Fred's house being hot (*FH*) when Fred's air conditioner is working (*FB* is false), in which case the outside (*OT*) is not relevant. And similarly for Mary's house. See Figure 7. What is important to note is that *FH* and *MH* are dependent, but only if both air conditioners are broken, in which case the thermostat settings are irrelevant.

Suppose we were to sum out *OT* in VE. Once *OT* is eliminated, *FH* and *MH* become dependent. In VE and bucket elimination we form a factor $f(FH, MH, FT, FB, MB, MT, S)$ containing all the remaining variables. This factor represents $P(FH, MH|FT, FB, MB, MT, S)$ (unless there is a pathological case such as if *MT* or *MB* is a descendent of *FH*, or if *FT* or *FB* is a descendent of *MH*). One could imagine a version of VE that builds a tree-based representation for this factor. We show here how the confactor-based version is exploiting more structure than this.

If we are to take the contextual independence into account, we need to consider the dependence between *FH* and *MH* when both *FB* and *MB* are true (in which case *FT* and *MT* are irrelevant). For all of the other contexts, we can treat *FH* and *MH* as independent. The algorithm CVE does this automatically.

The conditional probabilities of Figures 6 and 7 can be represented as the following confactors:

$$\left\langle fb, \begin{array}{|cc|c|} \hline FH & OT & Value \\ \hline true & true & p_1 \\ true & false & p_2 \\ false & true & 1 - p_1 \\ false & false & 1 - p_2 \\ \hline \end{array} \right\rangle \tag{33}$$

$$\left\langle \overline{fb}, \begin{array}{|cc|c|} \hline FH & FT & Value \\ \hline true & true & p_3 \\ true & false & p_4 \\ false & true & 1 - p_3 \\ false & false & 1 - p_4 \\ \hline \end{array} \right\rangle \tag{34}$$

$$\left\langle mb, \begin{array}{|cc|c|} \hline MH & OT & Value \\ \hline true & true & p_5 \\ true & false & p_6 \\ false & true & 1 - p_5 \\ false & false & 1 - p_6 \\ \hline \end{array} \right\rangle \tag{35}$$

$$\left\langle \overline{mb}, \begin{array}{|cc|c|} \hline MH & MT & Value \\ \hline true & true & p_7 \\ true & false & p_8 \\ false & true & 1 - p_7 \\ false & false & 1 - p_8 \\ \hline \end{array} \right\rangle \tag{36}$$

$$\begin{array}{|cc|c|} \hline OT & S & Value \\ \hline true & true & p_9 \\ true & false & p_{10} \\ false & true & 1 - p_9 \\ false & false & 1 - p_{10} \\ \hline \end{array} \tag{37}$$

Eliminating *OT* from these confactors results in six confactors:

$$\left\langle fb \wedge mb, \begin{array}{|ccc|c|} \hline FH & MH & S & Value \\ \hline true & true & true & p_9 * (p_5 * p_1) + (1 - p_9) * (p_6 * p_2) \\ true & true & false & p_{10} * (p_5 * p_1) + (1 - p_{10}) * (p_6 * p_2) \\ true & mbalse & true & p_9 * ((1 - p_5) * p_1) + (1 - p_9) * ((1 - p_6) * p_2) \\ true & mbalse & false & p_{10} * ((1 - p_5) * p_1) + (1 - p_{10}) * ((1 - p_6) * p_2) \\ false & true & true & p_9 * (p_5 * (1 - p_1)) + (1 - p_9) * (p_6 * (1 - p_2)) \\ false & true & false & p_{10} * (p_5 * (1 - p_1)) + (1 - p_{10}) * (p_6 * (1 - p_2)) \\ false & false & true & p_9 * ((1 - p_5) * (1 - p_1)) + (1 - p_9) * ((1 - p_6) * (1 - p_2)) \\ false & false & false & p_{10} * ((1 - p_5) * (1 - p_1)) + (1 - p_{10}) * ((1 - p_6) * (1 - p_2)) \\ \hline \end{array} \right\rangle$$

$$\left\langle fb \wedge \overline{mb}, \begin{array}{|ll|l|} \hline FH & S & Value \\ \hline true & true & p_9 * p_1 + (1 - p_9) * p_2 \\ true & false & p_{10} * p_1 + (1 - p_{10}) * p_2 \\ false & true & p_9 * (1 - p_1) + (1 - p_9) * (1 - p_2) \\ false & false & p_{10} * (1 - p_1) + (1 - p_{10}) * (1 - p_2) \\ \hline \end{array} \right\rangle$$

$$\left\langle \overline{fb}, \begin{array}{|ll|l|} \hline FH & FT & Value \\ \hline true & true & p_3 \\ true & false & p_4 \\ false & true & 1 - p_3 \\ false & false & 1 - p_4 \\ \hline \end{array} \right\rangle$$

$$\left\langle \overline{fb} \wedge mb, \begin{array}{|ll|l|} \hline MH & S & Value \\ \hline true & true & p_9 * p_5 + (1 - p_9) * p_6 \\ true & false & p_{10} * p_5 + (1 - p_{10}) * p_6 \\ false & true & p_9 * (1 - p_5) + (1 - p_9) * (1 - p_6) \\ false & false & p_{10} * (1 - p_5) + (1 - p_{10}) * (1 - p_6) \\ \hline \end{array} \right\rangle$$

$$\left\langle \overline{fb} \wedge \overline{mb}, \begin{array}{|l|l|} \hline S & Value \\ \hline true & p_9 + (1 - p_9) \\ false & p_{10} + (1 - p_{10}) \\ \hline \end{array} \right\rangle$$

$$\left\langle \overline{mb}, \begin{array}{|ll|l|} \hline MH & MT & Value \\ \hline true & true & p_7 \\ true & false & p_8 \\ false & true & 1 - p_7 \\ false & false & 1 - p_8 \\ \hline \end{array} \right\rangle$$

Note that the third and the sixth confactors were there originally and were not affected by eliminating *OT*.

The resultant confactors encode the probabilities of {*FH*, *MH*} in the context $fb \wedge mb$. For all other contexts, CVE considers *FH* and *MH* separately. The total table size of the confactors after *OT* is eliminated is 24.

Unlike VE or BEBA, we need the combined effect on *FH* and *MH* only for the contexts where *OT* is relevant to both *FH* and *MH*. For all other contexts, we don't need to combine the confactors for *FH* and *MH*. This is important, as combining the confactors is the primary source of combinatorial explosion. By avoiding combining confactors, we can have a potentially huge saving when the variable to be summed out appears in few contexts.

### 4.13 CVE Compared To VE

It is interesting to see the relationship between the confactors generated and the factors of VE for the same belief network, with the same query and observations and the same elimination ordering. There are two aspects to the comparison, first the exploitation of contexts, and second the idea of not multiplying confactors unless you need to. In this section, we introduce a tree-based variable elimination algorithm (TVE) that uses confactors but doesn't have the property of the example above where there are confactors that are not multiplied when VE would multiply the corresponding factors.

In order to understand the relationship between VE and CVE for the same query and the same elimination order, we can consider the VE derivation tree of the final answer. The tree contains all initial and intermediate factors created in VE. The parents in the tree of any factor are those factors that were multiplied or had a variable summed out to produce this factor. Note that this is a tree (each factor has only one child) as each factor is only used once in VE; once it is multiplied or a variable is summed from it, it is removed.

For each node in this tree that is created by multiplying two other factors, the number of multiplications in VE is equal to the table size of the resulting factor. For each factor created by summing out a variable, the number of additions is equal to the size of its parent minus its size.

We can define tree-based variable elimination (TVE) to be a composite of VE and CVE. It uses confactors as in CVE. Associated with each factor in the VE derivation tree is a set of confactors. When VE multiplies two factors, TVE multiplies (and does the requisite splitting) all of the compatible confactors associated with the factors being multiplied. TVE is essentially the same as the tree-based merging of Boutilier (1997) (but Boutilier also does maximization at decisions).

Whenever VE multiplies two factors, TVE multiplies all of the confactors associated with the factors. The TVE confactors associated with the VE factors will always have a total table size that is less than or equal to the VE factor size. TVE maintains a set of confactors with mutually exclusive and covering contexts. The number of multiplications is equal to the resulting table size for each pairwise multiplication (as each entry is computed by multiplying two numbers). It is easy to see that TVE always does fewer or an equal number of multiplications than VE.

CVE is like TVE except that CVE doesn't multiply some of the confactors when VE multiplies two factors. It delays the multiplications until they need to be done. It relies on the hope that the confactors can be separately simplified before they need to be multiplied. This hope is not unjustified because if eliminating a variable means that both of the factors need to be multiplied by other confactors, then they need to be multiplied by each other.

**Example 21** If we were to use TVE for Example 20, once $OT$ is eliminated, TVE builds a tree representing the probability on both $FH$ and $MH$. This entails multiplying out the confactors that were not combined in CVE, for example multiplying the third, fifth and sixth factors of the result of Example 20, which produces the confactor of the form $\langle \overline{fb} \wedge \overline{mb}, t(FH, FT, MH, MT, S) \rangle$. Eliminating $OT$ results in a set of confactors with total table size 72, compared to 24 for VE. Without any contextual structure, VE builds a table with $2^7 = 128$ values.

It is possible that not multiplying compatible confactors earlier means that we will eventually have to do more multiplications. The following example is the simplest example we could find where CVE does more multiplications than VE or TVE. Slight variations in the structure of this example, however result in CVE doing fewer multiplications.

**Example 22** Consider the belief network shown in Figure 8(a). First we will sum out a variable, $A$, to create two confactors that are not multiplied in CVE but are multiplied in TVE. We then multiply one of these confactors by another factor when summing out the second variable, $B$. We then force the multiplication when eliminating $C$.

Suppose that all of the variables are binary except for variable $W$ that has domain size 1000. (The counter example doesn't rely on non-binary variables; you could just have $B$ having 10 binary parents, but this makes the arithmetic less clear). In the analysis below we only discuss the multiplications that involve $W$, as the other multiplications sum up to less than a hundred, and are dominated by the multiplications that involve $W$.

Suppose that we have the following confactors for $S$:

$$\langle x, t_1(A, B, C, S) \rangle \tag{38}$$

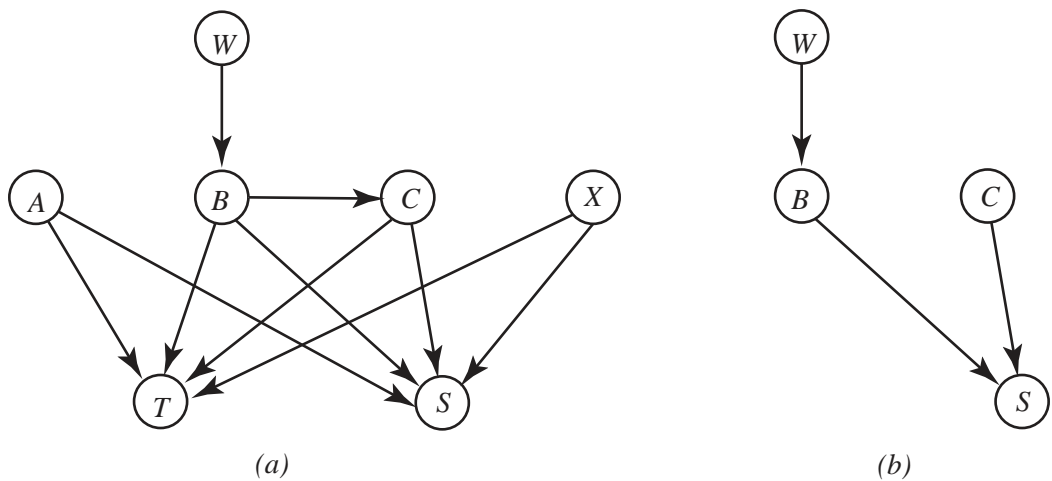

Figure 8: Belief network for Example 22.

$$\langle \bar{x}, t_2(B, C, S) \rangle \tag{39}$$

Suppose that we have the following confactors for $T$:

$$\langle x, t_3(A, B, C, T) \rangle \tag{40}$$

$$\langle \bar{x}, t_4(C, T) \rangle \tag{41}$$

Suppose first that $T$ is observed. Then the confactors for $T$ are replaced by:

$$\langle x, t_5(A, B, C) \rangle \tag{42}$$

$$\langle \bar{x}, t_6(C) \rangle \tag{43}$$

Let's now eliminate $A$. In both CVE and TVE, confactors (38) and (42) and the prior on $A$ are multiplied together, and $A$ is summed out resulting in:

$$\langle x, t_7(B, C, S) \rangle \tag{44}$$

In *TVE*, confactors (39) and (43) are also multiplied together resulting in:

$$\langle \bar{x}, t_8(B, C, S) \rangle \tag{45}$$

Next we eliminate $B$. In both CVE and TVE, confactor (44) is multiplied by the factors representing $P(C|B)$ and $P(B|W)$. We assume that the factor representing $P(B|W)$ is multiplied last as this minimises the number of multiplications. This involves 8008 multiplications (as the product has a table size of 8000 and the intermediate table is of size 8). Then $B$ is summed out resulting in the factor:

$$\langle x, t_9(C, S, W) \rangle \tag{46}$$

In CVE, confactor (39) is multiplied by the factors representing $P(C|B)$ and $P(B|W)$. This also involves 8008 multiplications. Then $B$ is summed out from the product resulting in:

$$\langle \bar{x}, t_{10}(C, S, W) \rangle \tag{47}$$

In TVE, confactor (45) is multiplied by the factors representing $P(C|B)$ and $P(B|W)$. This also involves 8008 multiplications. Then $B$ is summed out from the product resulting in:

$$\langle \bar{x}, t_{11}(C, S, W) \rangle \tag{48}$$

When $C$ is eliminated, TVE requires no more multiplications. It just sums out $C$ from the table of confactor (48). However in CVE, we need to multiply confactors (43) and (47), which involves 4000 multiplications. The resulting confactors from CVE and TVE are identical.

Thus CVE requires about 20000 multiplications, TVE requires about 16000 multiplications. VE, for the same elimination order, requires about 16000 multiplications.

This should not be surprising, particularly when you realise that VE is not optimal. For a given elimination ordering, it is sometimes optimal to multiply factors before VE actually multiplies them, as the following example shows:

**Example 23** Consider the belief network in Figure 8(b), with the same domain sizes as in the previous example. The factors represent $P(W)$, $P(B|W)$, $P(C)$, $P(S|BC)$. If we were to eliminate $B$ then $C$, it is more efficient to preemptively multiply $P(C)$ by $P(S|BC)$ than to delay the multiplication till after summing out $B$ (as VE does).

It may seem negative to show that CVE doesn't always do fewer multiplications than VE but has the overhead of maintaining contexts. However, there seems no reason why the preemptive multiplication of TVE is optimal either. One possibility is to treat "when to multiply" and "when to sum out variables" as a secondary optimisation problem (Bertelè and Brioschi, 1972; Shachter et al., 1990; D'Ambrosio, 1995); unfortunately this optimization is also computationally difficult (D'Ambrosio, 1995). This, however is beyond the scope of this paper.

## 5. Avoiding Splitting

### 5.1 Absorption

In this section we characterize a case when we don't need to split during multiplication. Note that the result of eliminating a variable is exactly the same as before; we are saving because we don't create the residuals, but rather use the original confactor.

**Definition 15** A multiset of confactors $R$ is **complete** if the contexts of the confactors are mutually exclusive and covering.

If we have a multiset $R$ of confactors that is complete, it means that we don't have to split other confactors $r$ that may need to be multiplied by all of the confactors in $R$. For each residual of $r$, there will be another element of $R$ with which it will be compatible. Instead of splitting $r$, we can just multiply it by each element of $R$. This is the intuition behind absorption. Note that we may need to split elements of $R$ during multiplication.

Suppose we have a complete multiset of confactors $R$ and another confactor $r_1 = \langle c_1, t_1 \rangle$. Define

$$absorb(R, \langle c_1, t_1 \rangle)$$
$$= \{\langle c_i, p_i \rangle \in R : incompatible(c_i, c_1)\}$$
$$\cup \bigcup_{\substack{\langle c_i, t_i \rangle \in R \\ compatible(c_1, c_i)}} (residual(\langle c_i, t_i \rangle, c_1) \cup \{\langle c_1 \cup c_i, set(t_1, c_i) \otimes_t set(t_i, c_1) \rangle\})$$

where $\otimes_t$ is table multiplication (Definition 3).

**Correctness:** We can replace $R \cup \{r_1\}$ with $absorb(R, r_1)$ and the program invariant is preserved. First note that the confactors in $absorb(R, r_1)$ are complete (and so the second part of the invariant holds). To see why the first part of the invariant is preserved, consider a complete context $c$. If $c$ is compatible with $c_1$, then in the original confactors, we use one confactor from $R$ as well as $r_1$. In the revised confactor base we use the appropriate confactor with the product of the original probabilities. If $c$ is incompatible with $c_1$ then it is compatible with one element $\langle c_2, p_2 \rangle$ of $R$. If $c_2$ is incompatible with $c_1$, the confactor to be used is the original confactor, and if $c_2$ is compatible with $c_1$, then we use the residual confactor. In each case we get the same contributions from $R \cup \{r_1\}$ and from $absorb(R, r_1)$.

**Example 24** If we were to eliminate $B$ from the confactors of Figure 4 (as in example 14), we can treat the two confactors for $P(B|Y, Z)$ as the complete multiset of confactors. This means that we don't need to split the other confactors on $Y$.

Note that if we try to use the confactors for $E$ as the complete set of confactors when eliminating $B$, we don't have to split the confactors on $A$, $C$ or $D$, but we have to consider confactors that don't involve $B$ when eliminating $B$, and we end up multiplying confactors that don't need to be multiplied.

Note that if $R$ cannot be represented as a decision tree (e.g., if $R$ has confactors corresponding to the contexts in $\{\{a, b\}, \{\overline{a}, c\}, \{\overline{b}, \overline{c}\}, \{a, \overline{b}, c\}, \{\overline{a}, b, \overline{c}\}\}$), it's possible that there is no way to split $r_1$ that results in as compact a representation as results from $absorb(R, r_1)$.

It seems as though it is very useful to have a multiset of confactors that is complete, but it is not of much use if we cannot easily find such a set. First note that if we have a multiset $R$ of confactors that is complete, then $absorb(R, r_1)$ is also a multiset of confactors that is complete, which can, in turn, be used to combine with another confactor.

Initially, for each variable $X$, the confactors that represent the conditional probability table $P(X|\pi_X)$ are complete. Moreover they will all contain $X$ and so need to be involved when eliminating $X$. These can be used as the initial seed for absorbing other confactors. Unfortunately, after we have eliminated some variables, the confactors that define the initial conditional probability tables for some variable $X$ don't exist anymore; they have been split, multiplied by other confactors and added to other confactors. However, for each variable $X$, we can still extract a useful multiset of confactors that all contain $X$ that are complete on the empty context (i.e., are mutually exclusive and covering). These will be called the **confactors for** $X$, and correspond to the confactors with $X$ in the head in an earlier version of contextual variable elimination (Poole, 1997).

**Definition 16** If $X$ is a variable, the **confactors for** $X$ are a subset of the confactor base defined by:

- Initially the confactors for $X$ are the confactors that define the conditional probability $P(X|\pi_X)$.

- When we split a confactor for $X$, the confactors created are also confactors for $X$. Note that this does not cover the case where we are splitting a confactor on a confactor for $X$.

- When we multiply a confactor for $X$ with another confactor, the confactor created is a confactor for $X$.

- When we add a confactor for $X$ with another confactor (when eliminating another variable $Y$, for example), the resulting confactor is also a confactor for $X$.

**Proposition 2** The confactors for $X$ at any stage of contextual variable elimination are complete.

To show this, we will show that the three basic operations preserve this property.

- splitting preserves this property, as the resulting confactors are exclusive and cover the context of the confactor being split.

- multiplication preserves this property as, for any variable $X$, only one of the confactors involved in a multiplication can be a confactor for $X$ (as the confactors for $X$ are mutually exclusive) and the set of contexts covered by the confactors isn't changed by multiplication. This argument also holds for absorption.

- for addition, note that, for any variable $X$, either all of the confactors or none of the confactors involved in an addition are confactors for $X$ when summing out $Y$. To show this suppose we have $r_1 = \langle c \wedge Y = v_1, p_1 \rangle$ and $r_2 = \langle c \wedge Y = v_2, p_2 \rangle$ where confactor $r_1$ is a confactor for $X$ and confactor $r_2$ isn't. Because the confactors for $X$ are mutually exclusive and covering, there must be a confactor that is covered by $X$ that is applicable in a context when $c \wedge Y = v_2$ is applicable. This confactor cannot mention $Y$, for otherwise addition isn't applicable, and so it must also be applicable when $c \wedge y = v_1$ is true, which contradicts the mutual exclusiveness of the confactors for $X$. Now it is easy to see that addition preserves this property as the confactors being summed cover the same contexts as the resulting confactor.

We can also use the idea of the confactors for $X$ to recognise when summing out a variable will create a table of 1's that can be removed (see Section 4.10). First note that in the original confactors for $X$, if, when eliminating $X$, we don't have to multiply these by other confactors (i.e., they have no children in this context), then we know that summing out $X$ will produce a table of 1's. We can do better than this for recognising when we will produce ones. We will use one bit of information to encode whether a confactor for $X$ is *pure* for $X$. Initially all of the confactors for $X$ are pure for $X$. If a confactor for $X$ is pure for $X$, and, when eliminating $Y$ is absorbed into a confactors for $Y$ that is pure for $Y$, then the resulting confactor is pure for $X$. For every other case when a confactor for $X$ is multiplied by another confactor, the result is not pure for $X$. If we are summing out $X$, after absorption, we can remove all confactors for $X$ that are pure for $X$. This is correct because we have maintained the invariant that if we sum out $X$ from the table of a confactor for $X$ that is pure for $X$ we create a table with only ones. Note that this procedure generalises the idea that we can recursively remove variables with no children that are neither observed nor queried.

The idea of absorption into the rules for the variable being eliminated described in this section should only be seen as a heuristic to avoid splitting. It does not necessarily reduce the amount of work. First note that in variable elimination there is a choice in the order that we multiply factors. Multiplication of factors is associative and commutative, however, the order in which the multiplication is carried out affects efficiency, as the following example shows.

**Example 25** Suppose variable $B$ has one parent, $A$ and two children $C$ and $D$, who each only have $B$ as a parent. When eliminating $B$ we have to multiply the factors representing $P(B|A)$, $P(C|B)$ and $P(D|B)$. Suppose that $B$, $C$ and $D$ are binary, and that $A$ has domain size of 1000. When multiplying two factors the number of multiplications required is equal to the size of the resulting factor. If we save intermediate results, and multiply these in the order $(P(B|A) \otimes_t P(C|B)) \otimes_t P(D|B)$, we will do 12000 multiplications. If we save intermediate results, and multiply these in the order $P(B|A) \otimes_t (P(C|B) \otimes_t P(D|B))$, we will do 8008 multiplications. If we don't save intermediate tables, but instead recompute every product, we will do 16000 multiplications.

If you need to multiply $k > 1$ factors, where $m$ is the size of the resulting factor, the number of multiplications is bounded below by $k - 2 + m$ (as the final product requires $m$ multiplications and each other requires at least one multiplication) and bounded above by $(k - 1) * m$ (as there are $k - 1$ factor multiplications and each of these requires at most $m$ multiplications).

The associative ordering imposed by absorption into the rules for the variable being eliminated (which for the example above implies absorbing $P(C|B)$ and $P(D|B)$ into $P(B|A)$) may not be the optimal multiplication ordering. The absorption ordering (that saves because it reduced splitting) should be seen as a heuristic; it may be worthwhile to do a meta-level analysis to determine what order to multiply (Bertelè and Brioschi, 1972; Shachter et al., 1990; D'Ambrosio, 1995), but this is beyond the scope of this paper.

### 5.2 Summing Out A Variable

Suppose we are eliminating $Y$, and we have absorbed all of the confactors that contain $Y$ into the confactors for $Y$. Then any two confactors in $R$ that contain $Y$ have incompatible contexts. The contexts for the confactors that contain $Y$ in the table are disjoint from the contexts of the confactors that contain $Y$ in the body.

Summing out $Y$ from a confactor that contains $Y$ in the table proceeds as before. We can use a similar trick to absorption to avoid any more splitting when adding confactors that contain $Y$ in the body.

Suppose $Y$ has domain $\{v_1, \ldots, v_s\}$. The contexts of the confactors with $Y{=}v_i$ in the body are exclusive and the disjunct is logically equivalent to the disjunct of confactors with $Y{=}v_j$ in the body for any other value $v_j$.

For each $1 \leq i \leq s$, let $R_i = \{\langle b, t\rangle : \langle b \wedge Y{=}v_i, t\rangle \in R^+\}$. Thus $R_i$ is the confactor for $Y = v_i$ in the context, but with $Y = v_i$ omitted from the context. We will combine the $R_i$'s using a binary operation:

$$R_1 \oplus_g R_2 = \{\langle c_1 \cup c_2, set(t_1, c_2) \oplus_t set(t_2, c_1)\rangle : \langle c_1, t_1\rangle \in R_1, \langle c_2, t_2\rangle \in R_2, compatible(c_1, c_2)\}$$

where $\oplus_t$ is an addition operation defined on tables that is identical to the product $\otimes_t$ of Definition 3 except that it adds the values instead of multiplying them.

Then $R_1 \oplus_g R_2 \oplus_g \ldots \oplus_g R_s$ is the result from summing out $Y$ from the confactors with $Y$ in the body.

### 5.3 Contextual Variable Elimination with Absorption

A version of *contextual variable elimination* that uses absorption, is given in Figure 9. This is the algorithm used in the experimental results of Section 6.

The elimination procedure is called once for each non-query, non-observed variable. The order in which the variables are selected is called the **elimination ordering**. This algorithm does not imply that the elimination ordering has to be given a priori.

One of the main issues in implementing this algorithm is efficient indexing for the confactors. We want to be able to quickly find the confactors for $Y$, the confactors that contain $Y$, and the compatible confactors during addition and absorption. If we are given a prior elimination ordering, we can use the idea of bucket elimination (Dechter, 1996), namely that a confactor can be placed in the bucket of the earliest variable in the elimination ordering. When we eliminate $Y$, all of the confactors that contain $Y$ are in $Y$'s bucket. If we don't have a prior elimination ordering, we can keep an inverted

**To compute** $P(X|E_1{=}o_1 \wedge \ldots \wedge E_s{=}o_s)$
    **Given**    multiset of contextual contribution confactors
    1.    Incorporate evidence as in Section 4.7.
    2.    While there is a factor involving a non-query variable
              Select non-query variable $Y$ to eliminate;
              Call *eliminate*$(Y)$.
    3.    Compute posterior probability for $X$ as in Section 4.8

**Procedure** *eliminate*$(Y)$:
    **partition** the confactorbase $R$ into:
        $R^-$ those confactors that don't involve $Y$
        $R^+ = \{r \in R : r \text{ is a confactor for } Y\}$
        $R^* = \{r \in R : r \text{ involves } Y \text{ and } r \text{ is not a confactor for } Y\}$;
    **for each** $r \in R^*$
        **do** $R^+ \leftarrow absorb(R^+, r)$;
    **partition** $R^+$ into:
        $R^t = \{r \in R^+ : Y \text{ in table of } r\}$
        $R_i = \{\langle b, t \rangle : \langle b \wedge Y{=}v_i, t \rangle \in R^+\}$ for each $1 \leq i \leq s$, where $dom(Y) = \{v_1, \ldots, v_s\}$.
    **Return** confactorbase $R^- \cup (R_1 \oplus_g \cdots \oplus_g R_s) \cup \{\langle b_i, \sum_Y t_i \rangle : \langle b_i, t_i \rangle \in R^t\}$.

$absorb(R, r)$ is defined in Section 5.1.
$R_1 \oplus_g R_2$ is defined in Section 5.2.

Figure 9: Contextual Variable Elimination with Absorption

list of the confactors (for each variable, we have a list of all of the confactors that are for that variable and a list of the confactors that contain that variable). We then have to maintain these lists as we create new confactors and delete old ones. We also want to be able to index the confactors so that we can quickly find other confactors that contain the variable to be eliminated and have compatible contexts. In our implementation, we compared all of the confactors that contain the variable to be eliminated, and rejected those with incompatible contexts. Ideally, we should be able to do better than this, but how to do it is an open question.

There are a number of choice points in this algorithm:

- the elimination ordering.

- the splitting ordering; when computing residuals, which order should the variables be split on. This is discussed in Section 4.4.

- the order that the elements of $R^*$ are absorbed. This has an impact similar to the choice of multiplication ordering for VE (when we have to multiply a number of factors, which order should they be done); sometimes we can have smaller intermediate factors if the choice is done appropriately.

## 6. Empirical Results

An interesting question is whether there are real examples where the advantage of exploiting contextual independence outweighs the overhead of maintaining confactors.

We can easily generate synthetic examples where VE is exponentially worse than contextual variable elimination (for example, consider a single variable with $n$, otherwise unconnected, parents, where the decision tree for the variable only has one instance of each parent variable, and we eliminate from the leaves). At another extreme, where all contexts are empty, we get VE with very little overhead. However, if there is a little bit of CSI, it is possible that we need to have the overhead of reasoning about variables in the contexts, but get no additional savings. The role of the empirical results is to investigate whether it is ever worthwhile trying to exploit context-specific independence, and what features of the problem lead to more efficient inference.

### 6.1 A Pseudo-Natural Example

While it may seem that we should be able to test whether CVE is worthwhile for natural examples by comparing it to VE for standard examples, it isn't obvious that this is meaningful. With the table-based representations, there is a huge overhead for adding a new parent to a variable, however there is no overhead for making a complex function for how a variable depends on its existing parents. Thus, without the availability of effective algorithms that exploit contextual independence where there is a small overhead for adding a variable to restricted contexts, it is arguable that builders of models will tend to be reluctant to add variables, but will tend to overfit the function for how a variable depends on its parents. As all models are approximations it makes sense to consider approximations to standard models. As we are not testing the approximations (Dearden and Boutilier, 1997; Poole, 1998), we will pretend these are the real models.

In this section we produce evidence that there exists networks for which CVE is better than VE. The sole purpose of this experiment it to demonstrate that there potentially are problems where it is worthwhile using CVE. We use an instance of the *water* network (Jensen, Kjærulff, Olesen and Pedersen, 1989) from the Bayesian network repository[8] where we approximated the conditional probabilities to create contextual independencies. Full details of how the examples were constructed are in Appendix A. We collapsed probabilities that were within 0.05 of each other to create confactors. The *water* network has 32 variables and the tabular representation has a table size of 11018 (after removing variables from tables that made a difference of less that 0.05). The contextual belief network representation we constructed had 41 confactors and a total table size of 5834.

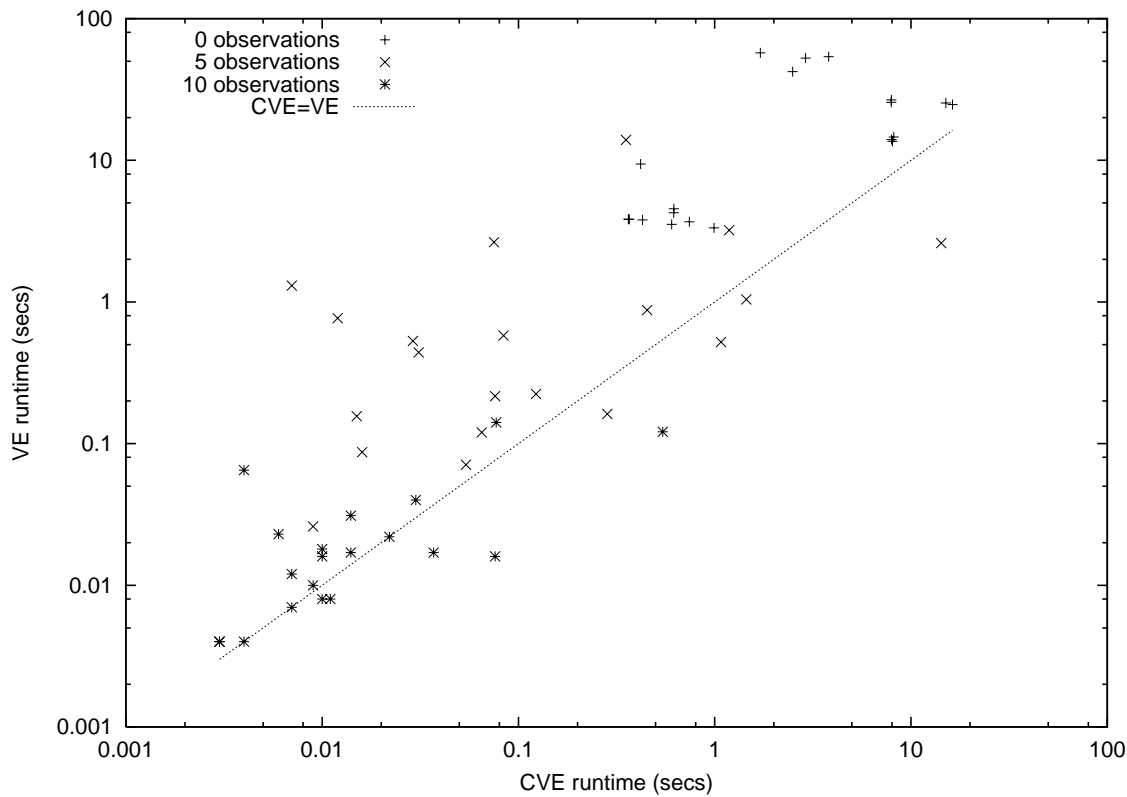

Figure 10: Scatterplot of runtimes (in msecs) of CVE (x-axis) and VE (y-axis) for the *water* network. Full details are in Appendix A.

Figure 10 shows a scatter plot of 60 runs of random queries[9]. There were 20 runs each for 0, 5 and 10 observed variables. The raw data is presented in Appendix A. The first thing to notice is that, as the number of observations increases, inference becomes much faster. CVE was often significantly faster than VE. There are a few cases where CVE was much worse than VE; essentially, given the elimination ordering, the context-specific independence didn't save us anything in these example, but we had to pay the overhead of having variables in the context.

---

8. http://www.cs.huji.ac.il/labs/compbio/Repository/
9. Note that all of these results are statistically significant to some degree. The least significant is with 10 observations, that, using the matched-sample t-test (also known as the paired t-test), is significant to the 0.2 level for the logarithm of the runtimes. The others are significant way below the 0.001 level. The logarithm is appropriate as the difference in the logarithms corresponds to the multiplicative speedup.

### 6.2 Randomised Networks

In order to see how the algorithm depends on the structure inherent in the network, we constructed a number of parametrized classes of networks. We explicitly constructed networks that display contextual independence, as if there is not contextual independence this algorithm degenerates to VE.

We have the following parameters for building random networks:

$n$ the number of variables

$s$ the number of splits (so there will be $n + s$ confactors).

$p$ the probability that a predecessor variable that isn't in the context of a confactor will be in the table of the confactor.

The exact algorithm for constructing the examples is given in Appendix A.

The variable $s$ controls the number of confactors, and $p$ controls (probabilistically) the size of the tables. Figure 11[10] shows a scatter plot comparing the runtimes of CVE and VE for $n = 30$ and $p = 0.2$ and for three different values of $s$, 5, 10, and 15.

While this may look reasonable, it should be noticed that the number of splits and the number of different variables in the splits is strongly correlated in these examples (see Appendix A for details). However, one of the properties of CVE is that if a variable does not appear in the body of any confactor, it is never added to the context of a constructed confactor. That is, a variable that only appears in tables, always stays in tables. Thus it may be conjectured that having fewer variables appearing in contexts may be good for efficiency.

We carried out another experiment to test this hypothesis. In this experiment, the networks were generated as before, however, when we went to split a context we attempted to first split it using a variable that appears in a different context before using a variable that didn't appear in a context. The full details of the algorithms to generate examples and some example data are given in Appendix A. Figure 12 shows a scatter plot of comparing the run times for CVE and VE for each of the generated examples. With this class of networks CVE is significantly faster than VE.

## 7. Comparison With Other Proposals

In this section we compare CVE with other proposals for exploiting context-specific information.

The belief network with the conditional probability table of Figure 1 (i.e., with the contextual independence shown in Figure 4) is particularly illuminating because other algorithms do very badly on it. Under the elimination ordering $B, D, C, A, Y, Z$, to find the marginal on $E$, the most complicated confactor multiset created is the confactor multiset for $E$ after eliminating $B$ (see Example 14) with a total table size of 16. Observations simplify the algorithm as they mean fewer partial evaluations.

In contrast, VE requires a factor with table size 64 after $B$ is eliminated. Clique tree propagation constructs two cliques, one containing $Y, Z, A, B, C, D$ of size $2^6 = 64$, and the other containing $A, B, C, D, E$ of size 32. Neither takes the structure of the conditional probabilities into account.

Note however, that VE and clique tree propagation manipulate tables which can be indexed much faster than we can manipulate confactors. There are cases where the total size of the tables of the confactors is exponentially smaller than the tables (where added variables are only relevant in narrow contexts). There are other cases where the table size for the confactors is the same as the table size in VE, and where the overhead for manipulating contexts will not make CVE competitive with the table-based methods.

---

10. Note that all of these results are statistically significant. The least significant is the $s = 10$ plot that, using the matched-sample t-test (also known as the paired t-test), is significant to the 0.05 level for the logarithm of the runtimes. The logarithm is appropriate as the difference in the logarithms corresponds to the multiplicative speedup.

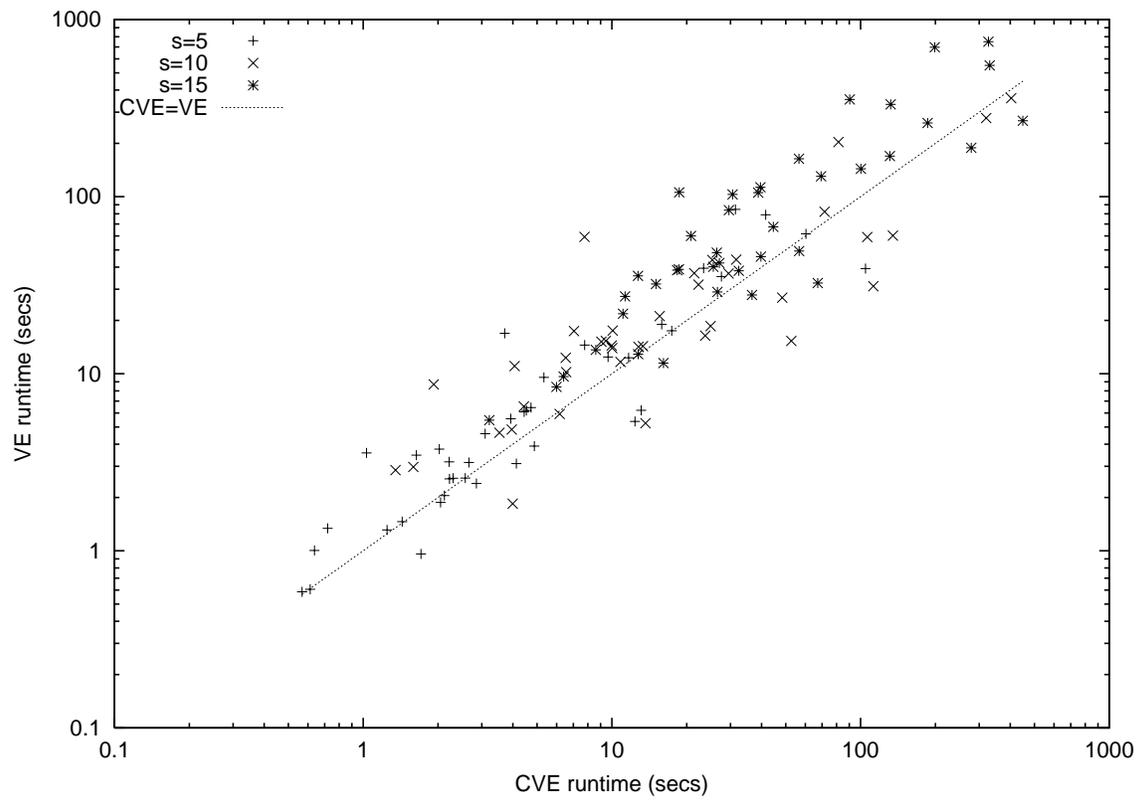

Figure 11: Scatterplot of runtimes (in seconds) of CVE (x-axis) and VE (y-axis) for randomised networks

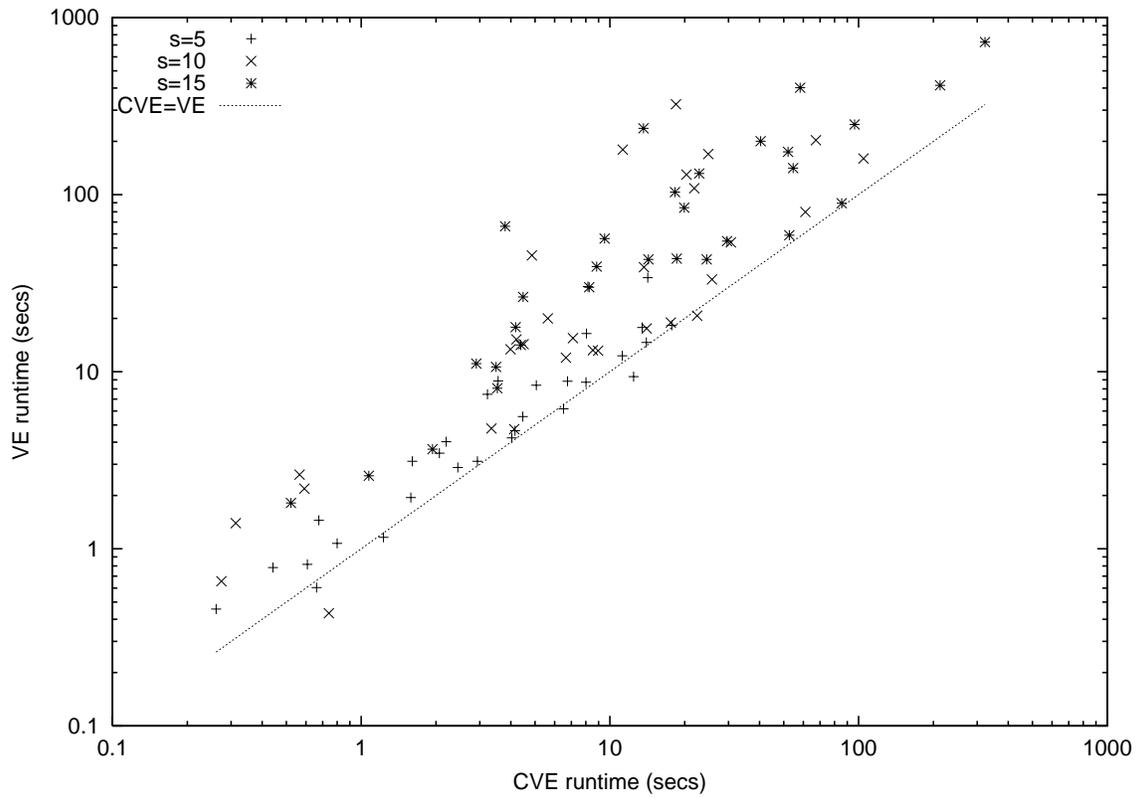

Figure 12: Scatterplot of runtimes (in seconds) of CVE (x-axis) and VE (y-axis) for random networks biased towards fewer different variables in the contexts.

Boutilier et al. (1996) present two algorithms to exploit structure. For the network transformation and clustering method, the example of Figure 1 is the worst case; no structure can be exploited after triangulation of the resulting graph. (The tree for $E$ in Figure 3 is structurally identical to the tree for $X(1)$ in Figure 2 of (Boutilier et al., 1996)). The structured cutset conditioning algorithm does well on this example. However, if the example is changed so that there are multiple (disconnected) copies of the same graph, the cutset conditioning algorithm is exponential in the number of copies, whereas the probabilistic partial evaluation algorithm is linear[11].

This algorithm is most closely related to the tree-based algorithms for solving MDPs (Boutilier et al., 1995), but these work with much more restricted networks and with stringent assumptions on what is observable. CVE is simpler than the analogous algorithm by Boutilier (1997) for structured MDP with correlated action effects that represents conditional probabilities as trees. Section 4.12 shows why we can do better than the tree-based algorithms.

Poole (1997) gave an earlier version of rule-based VE, but it is more complicated in that it distinguished between the head and the body of rules as part of the inference (although the confactors *for X* correspond to the rules with *X* in the head). CVE is much more efficient than the rule-based VE as it allows for faster indexing of tables. The notion of *absorption* was introduced by Zhang (1998), which motivated the work in a very different way. Zhang and Poole (1999) give a mathematical analysis of how context-specific independence can be exploited in terms of partial functions. The union-product is a formalization of the operation we are using between confactors. The current paper extends all of these in giving a specific algorithm showing how to combine the confactors and tables, gives a more general analysis of when we need to do confactor splitting and when we don't need to, gives a more sophisticated method to initialize absorption (maintaining the confactors for a variable) and gives a much more detailed empirical evaluation (with a new implementation). The ability to handle ones is also improved.

Finally the representation should be contrasted with that of Geiger and Heckerman (1996). They use a similarity network that gives a global decomposition of a belief network; for different contexts they allow for different belief networks. We allow for a local decomposition of conditional probabilities. The algorithms are not very similar.

## 8. Conclusion

This paper has presented a method for computing posterior probabilities in belief networks that exhibit context-specific independence. This algorithm is an instance of variable elimination, but when we sum out a variable the tables produced may depend on different sets of variables in different contexts.

The main open problem is in finding good heuristics for elimination orderings (Bertelè and Brioschi, 1972). Finding a good elimination ordering is related to finding good triangulations in building compact junction trees, for which there are good heuristics (Kjærulff, 1990; Becker and Geiger, 1996). These are not directly applicable to contextual variable elimination (although there are analogous heuristics that are applicable), as an important criteria in this case is the exact form of the confactors, and not just the graphical structure of the belief network. This means that it is not feasible to compute the elimination ordering a priori. We are also investigating anarchic orderings where we eliminate a variable in some contexts, without eliminating it in every context before we partially eliminate another variable. We believe that this opportunistic elimination of variables in contexts has much potential to improve efficiency without affecting correctness.

---

11. This does not mean that all conditioning algorithms need suffer from this. We conjecture that there is a conditioning algorithm that can extend contextual VE but save space, as in other bucket elimination algorithms (Dechter, 1996) or the relevant cutsets for probabilistic inference (Darwiche, 1995; Díez, 1996).

One of the main potential benefits of this algorithm is in approximation algorithms, where the confactors allow fine-grained control over distinctions. Complementary confactors with similar probabilities can be collapsed into a simpler confactor. This can potentially lead to more compact confactor bases, and reasonable posterior ranges (Dearden and Boutilier, 1997; Poole, 1998).

The work reported in this paper has been followed up by a number of researchers. Tung (2002) has shown how to exploit context-specific independence in clique trees. Guestrin, Venkataraman and Koller (2002) extended the contextual variable elimination to multi-agent coordination and planning.

## Acknowledgements

This work was supported by the Institute for Robotics and Intelligent Systems, Natural Sciences and Engineering Research Council of Canada Research Grant OGPOO44121 and Hong Kong Research Grants Council grant HKUST6093/99E. Thanks to Rita Sharma, Holger Hoos, Michael Horsch and the anonymous reviewers for valuable comments, and to Valerie McRae for careful proofreading. The code is available from the first author.

## Appendix A. Details of the experiments

### A.1 Water Network

In order to construct a pseudo-natural example, we used the *water* network from the Bayesian network repository[12] and modified it to let it exhibit context-specific independence. For each table, a variable was declared redundant if the differences in the probabilities for the values of this variable were less than a threshold of 0.05 from each other (thus, if we chose the midpoint of a reduced table, the original probabilities were all less than 0.025 from this midpoint). In order to discover context-specific independence, we carried out a greedy top-down algorithm to build a decision tree. If we are building the conditional for variable $X_i$, we chose the variable $Y$ to split on that results in the maximum number of pairs of numbers where the values for $X_i$ are within the threshold 0.05 of each other. We then recursively remove redundant variables from each side, and keep splitting. Once we have built a tree, for each node, we can decide whether to use the tabular representation or a factored representation. In these experiments, we only committed to the context-based representation when the total size of the context-based representation (obtained simply by summing the sizes of the tables) was less than 51% of the tabular representation.

It should be noted that these thresholds were not chosen arbitrarily. If we used 0.03 instead of 0.05, we didn't find any context-specific independence. If we chose 0.07 instead of 0.05, then the tabular representation collapses. If we chose 80% or 99% instead of 51% as the threshold for accepting a change, we got smaller tables, but much larger run times.

---

12. `http://www.cs.huji.ac.il/labs/compbio/Repository/`

|  | CVE |  | VE |  |
|---|---|---|---|---|
| Query Var | Time | Size | Time | Size |
| #11(CKND_12_15) | 15039 | 1327104 | 25368 | 5308416 |
| #13(CBODN_12_15) | 620 | 16384 | 4552 | 442368 |
| #27(CKND_12_45) | 3808 | 186624 | 53965 | 15925248 |
| #25(CKNI_12_45) | 1708 | 36864 | 57414 | 7077888 |
| #22(CKNN_12_30) | 367 | 16128 | 3821 | 442368 |
| #21(CBODN_12_30) | 7953 | 193536 | 13997 | 1769472 |
| #17(CKNI_12_30) | 742 | 48384 | 3677 | 442368 |
| #22(CKNN_12_30) | 363 | 16128 | 3846 | 442368 |
| #19(CKND_12_30) | 7939 | 774144 | 26654 | 2359296 |
| #15(CNON_12_15) | 8177 | 193536 | 14599 | 1769472 |
| #12(CNOD_12_15) | 618 | 37044 | 4264 | 442368 |
| #3(CKND_12_00) | 419 | 29376 | 9414 | 1327104 |
| #16(C_NI_12_30) | 429 | 28224 | 3799 | 442368 |
| #30(CKNN_12_45) | 2902 | 112896 | 52648 | 15925248 |
| #4(CNOD_12_00) | 2496 | 110592 | 42270 | 5308416 |
| #21(CBODN_12_30) | 8042 | 193536 | 13619 | 1769472 |
| #5(CBODN_12_00) | 992 | 112320 | 3334 | 442368 |
| #19(CKND_12_30) | 7936 | 774144 | 25637 | 2359296 |
| #11(CKND_12_15) | 16290 | 1327104 | 24753 | 5308416 |
| #23(CNON_12_30) | 604 | 37044 | 3535 | 442368 |

Figure 13: Data for random queries with no observed variables for the *water* network

Figure 13 shows some of the details of some of the data with no observations. Figures 14 and 15 shows some of the details of some of the data with 5 and 10 observation respectively. These make up the data shown in Figure 10 of the main paper. We show the index of the variable given the ordering in the repository (we started counting at 0).

All of the times are in milliseconds for a Java implementation running on 700 MHz Pentium running Linux with 768 megabytes of memory. In order to allow for the variation in run times due to garbage collection each evaluation was run three times and the smallest time was returned. The space is for the maximum table created in VE or the maximum size of the sum of all of the confactors created during the elimination of one variable.

## A.2 Randomised Experiments

The following procedure was used to generate the random networks. Given the $n$ variables, we impose a total ordering. We build a decision tree for each variable. The leaves of the decision trees correspond to contexts and a variable that the tree is for. We start with the empty decision tree for each variable. We randomly (with uniform probability) choose a leaf and a variable. If the variable is a possible split (i.e., it is a predecessor in the total ordering of the variable the leaf is for and the context corresponding to the leaf doesn't commit a value to that variable), we split that leaf on that variable. This is repeated until we have $n + s$ leaves. Then for each leaf, we built a confactor that has the same context and each predecessor of the variable that the leaf is for is included in the

|  |  | CVE | | VE | |
|---|---|---|---|---|---|
| Observed | Query Var | Time | Size | Time | Size |
| #1=2 #2=2 #26=2 #28=1 #30=1 | #8(C_NI_12_15) | 84 | 9216 | 579 | 36864 |
| #6=2 #8=1 #11=2 #14=2 #24=1 | #22(CKNN_12_30) | 15 | 816 | 156 | 9216 |
| #8=3 #14=0 #15=3 #18=3 #20=2 | #6(CKNN_12_00) | 9 | 336 | 26 | 2304 |
| #5=1 #9=0 #12=2 #15=0 #16=3 | #10(CBODD_12_15) | 54 | 576 | 71 | 6912 |
| #6=1 #7=0 #11=1 #13=3 #25=2 | #27(CKND_12_45) | 1184 | 28224 | 3210 | 147456 |
| #2=1 #6=0 #13=0 #19=1 #25=0 | #18(CBODD_12_30) | 123 | 9216 | 224 | 12288 |
| #0=3 #2=1 #18=2 #20=1 #23=1 | #17(CKNI_12_30) | 16 | 1728 | 87 | 5184 |
| #4=3 #10=3 #12=3 #17=0 #28=2 | #6(CKNN_12_00) | 284 | 6912 | 162 | 20736 |
| #3=1 #11=1 #19=1 #25=1 #31=3 | #14(CKNN_12_15) | 1450 | 36864 | 1041 | 147456 |
| #10=3 #19=0 #21=0 #26=0 #27=0 | #16(C_NI_12_30) | 1076 | 49152 | 521 | 49152 |
| #11=1 #13=0 #21=1 #25=2 #29=2 | #24(C_NI_12_45) | 353 | 28080 | 13928 | 1327104 |
| #3=0 #23=1 #26=1 #27=0 #28=3 | #15(CNON_12_15) | 14258 | 193536 | 2600 | 442368 |
| #9=2 #10=1 #13=1 #25=2 #26=1 | #30(CKNN_12_45) | 75 | 9216 | 2646 | 248832 |
| #9=1 #11=0 #14=2 #15=1 #27=0 | #25(CKNI_12_45) | 65 | 4096 | 120 | 4096 |
| #2=0 #10=2 #15=2 #17=1 #24=1 | #8(C_NI_12_15) | 12 | 576 | 767 | 110592 |
| #2=0 #7=3 #15=1 #21=2 #27=2 | #16(C_NI_12_30) | 29 | 1008 | 531 | 82944 |
| #0=3 #2=0 #6=0 #7=2 #16=2 | #29(CBODN_12_45) | 7 | 336 | 1304 | 147456 |
| #2=2 #20=2 #24=2 #25=0 #30=0 | #21(CBODN_12_30) | 453 | 49152 | 877 | 49152 |
| #8=2 #11=2 #19=2 #25=1 #29=3 | #9(CKNI_12_15) | 76 | 9216 | 216 | 9216 |
| #9=1 #12=3 #13=0 #19=2 #21=3 | #26(CBODD_12_45) | 31 | 1024 | 440 | 49152 |

Figure 14: Data for random queries with 5 randomly observed variables for the *water* network

|  |  | CVE | | VE | |
|---|---|---|---|---|---|
| Observed | Query | Time | Size | Time | Size |
| #2=3 #3=0 #7=2 #12=1 #13=2 #19=1 #21=2 #22=2 #26=1 #30=1 | #14 | 30 | 2304 | 40 | 2304 |
| #0=0 #3=0 #8=3 #12=3 #13=0 #16=2 #18=2 #26=3 #27=2 #28=1 | #7 | 14 | 256 | 17 | 768 |
| #1=2 #6=0 #8=0 #11=2 #17=0 #20=3 #22=1 #23=1 #24=2 #26=2 | #25 | 7 | 256 | 7 | 256 |
| #2=2 #5=3 #8=0 #9=1 #10=1 #17=2 #18=1 #22=2 #28=0 #30=1 | #23 | 4 | 192 | 4 | 192 |
| #4=2 #7=1 #8=2 #12=1 #13=2 #15=3 #17=0 #19=0 #22=2 #31=2 | #28 | 10 | 256 | 8 | 256 |
| #1=1 #7=1 #9=1 #10=0 #11=0 #13=3 #14=1 #23=1 #24=1 #30=2 | #22 | 9 | 336 | 10 | 768 |
| #2=1 #4=0 #6=0 #15=0 #18=0 #22=2 #23=3 #24=2 #29=2 #30=1 | #25 | 14 | 768 | 31 | 2304 |
| #3=0 #10=0 #14=1 #16=3 #19=0 #24=1 #25=2 #28=3 #30=1 #31=1 | #7 | 544 | 9216 | 121 | 9216 |
| #1=2 #2=3 #3=1 #9=2 #10=0 #14=0 #16=3 #25=1 #28=3 #30=2 | #8 | 10 | 1024 | 16 | 1024 |
| #2=3 #5=1 #6=0 #11=2 #12=0 #17=1 #22=0 #24=3 #27=0 #28=1 | #25 | 22 | 768 | 22 | 768 |
| #8=3 #9=2 #10=0 #11=1 #12=1 #14=2 #15=3 #19=0 #22=2 #26=3 | #5 | 3 | 84 | 4 | 192 |
| #1=0 #7=2 #8=2 #13=0 #15=3 #17=2 #20=3 #26=1 #27=0 #31=3 | #24 | 11 | 256 | 8 | 256 |
| #4=2 #5=3 #6=1 #9=1 #10=0 #12=2 #17=0 #19=1 #25=0 #29=0 | #23 | 6 | 256 | 23 | 1024 |
| #0=0 #2=1 #11=1 #13=1 #17=2 #21=3 #22=1 #23=1 #24=2 #30=2 | #27 | 10 | 576 | 18 | 768 |
| #4=1 #9=0 #10=0 #11=1 #12=2 #23=1 #25=2 #29=1 #30=0 #31=2 | #5 | 77 | 4096 | 141 | 12288 |
| #1=2 #6=0 #7=1 #10=3 #12=1 #15=3 #16=1 #17=2 #23=2 #24=1 | #27 | 3 | 64 | 4 | 144 |
| #1=2 #2=2 #3=2 #5=2 #9=2 #13=1 #15=0 #22=1 #25=2 #30=0 | #18 | 4 | 112 | 65 | 12288 |
| #0=3 #1=1 #5=1 #6=0 #7=1 #8=2 #15=3 #17=0 #24=3 #25=0 | #21 | 76 | 768 | 16 | 1024 |
| #0=2 #6=1 #8=0 #9=2 #10=2 #16=0 #18=0 #19=0 #21=0 #26=0 | #13 | 7 | 576 | 12 | 768 |
| #1=1 #3=0 #4=2 #9=1 #10=3 #13=0 #14=2 #22=0 #23=0 #30=2 | #6 | 37 | 768 | 17 | 768 |

Figure 15: Data for random queries with 10 randomly observed variables for the *water* network

*generate_random_CBN*(*n*, *s*, *p*) :
    create *n* Boolean variables $X_1, \ldots, X_n$
    Let $S = \{\langle \{\}, X_i \rangle : i \leq i \leq n\}$
        *{S is a set of context-variable pairs}*
    **repeat**
        choose random $\langle c, X_i \rangle \in S$
        choose random *j* such that $1 \leq j < n$
        **if** $j < i$ and $X_j$ doesn't appear in *c*:
            replace $\langle c, X_i \rangle \in S$ with $\langle c \wedge X_j = \textit{true}, X_i \rangle$ and $\langle c \wedge X_j = \textit{false}, X_i \rangle$
    **until** there are $n + s$ elements of *S*
    **for each** $\langle c, X_i \rangle \in S$
        Let $V = \{X_i\}$
        **for each** $j < i$
            **if** $X_j$ doesn't appear in *c*
                with probability *p* add $X_j$ to *V*
        create a random table *T* on the variables in *V*
        add confactor $\langle c, T \rangle$ to the contextual belief network

Figure 16: Algorithm for constructing the randomised examples

confactor with probability *p*. The variable that the leaf is for is also in the confactor. We assign random numbers the probabilities (these numbers won't affect anything in the results assuming that the times for operations on floating point numbers isn't affected by the actual values of the numbers).

The algorithm to construct the random examples used in Section 6.2 is given in Figure 16. Note that "choose random" means to choose randomly from a uniform distribution.

For the examples biased towards using fewer variables, we replaced the line:

    replace $\langle c, X_i \rangle \in S$ with $\langle c \wedge X_j = \textit{true}, X_i \rangle$ and $\langle c \wedge X_j = \textit{false}, X_i \rangle$

with

    **if** there is $k < i$ such that $X_k$ doesn't appear in *c* and $X_k$ is used in another context
        **then**
            replace $\langle c, X_i \rangle \in S$ with $\langle c \wedge X_k = \textit{true}, X_i \rangle$ and $\langle c \wedge X_k = \textit{false}, X_i \rangle$
        **else**
            replace $\langle c, X_i \rangle \in S$ with $\langle c \wedge X_j = \textit{true}, X_i \rangle$ and $\langle c \wedge X_j = \textit{false}, X_i \rangle$

where, if more than one such *k* exists, the *k* is chosen uniformly from all of the values that satisfy the condition.

Figure 17 shows some of the details of some of the data shown in Figure 11. All of the times are for a Java implementation running on a 700 MHz Pentium running Linux with 768 megabytes of memory. In order to allow for the variation in run times due to garbage collection, each evaluation was run three times and the smallest time was returned.

For each generated example, the table of Figure 17 shows

- *n* the number of variables

| n | s | p | CBN Size | BN Size | CVE Time | CVE MTS | VE Time | VE MTS |
|---|---|---|---|---|---|---|---|---|
| 30 | 5 (5) | 0.2 | 10922 | 8398266 | 31388 | 2097152 | 84801 | 4194304 |
| 30 | 5 (5) | 0.2 | 1846 | 132692 | 9653 | 393216 | 12418 | 524288 |
| 30 | 5 (5) | 0.2 | 1964 | 13456 | 2022 | 131072 | 3748 | 131072 |
| 30 | 5 (5) | 0.2 | 1600 | 4908 | 3922 | 196608 | 5572 | 524288 |
| 30 | 5 (5) | 0.2 | 1668 | 8292 | 60304 | 614400 | 61612 | 2097152 |
| 30 | 5 (5) | 0.2 | 904 | 1906 | 637 | 32768 | 1005 | 65536 |
| 30 | 5 (5) | 0.2 | 1738 | 10786 | 720 | 42048 | 1340 | 131072 |
| 30 | 5 (4) | 0.2 | 1744 | 18538 | 2223 | 131072 | 2546 | 262144 |
| 30 | 5 (4) | 0.2 | 3060 | 87292 | 11681 | 524288 | 12298 | 1048576 |
| 30 | 5 (3) | 0.2 | 2692 | 69602 | 5325 | 262144 | 9530 | 524288 |
| 30 | 10 (9) | 0.2 | 3842 | 530622 | 22288 | 524288 | 31835 | 2097152 |
| 30 | 10 (9) | 0.2 | 1262 | 36070 | 4063 | 147456 | 11038 | 524288 |
| 30 | 10 (10) | 0.2 | 3908 | 80704 | 112537 | 1966080 | 31214 | 4194304 |
| 30 | 10 (8) | 0.2 | 4904 | 33568176 | 81450 | 5111808 | 203284 | 16777216 |
| 30 | 10 (7) | 0.2 | 10456 | 314126 | 31627 | 589824 | 44079 | 2097152 |
| 30 | 10 (8) | 0.2 | 1790 | 28758 | 1590 | 98304 | 2974 | 262144 |
| 30 | 10 (9) | 0.2 | 2054 | 24452 | 13642 | 262144 | 5253 | 262144 |
| 30 | 10 (8) | 0.2 | 3608 | 58352 | 24948 | 819200 | 18574 | 1048576 |
| 30 | 10 (9) | 0.2 | 6392 | 1188654 | 403347 | 5767168 | 359992 | 33554432 |
| 30 | 10 (8) | 0.2 | 6180 | 42344 | 10078 | 253952 | 17501 | 1048576 |
| 30 | 15 (10) | 0.2 | 2724 | 2104338 | 56636 | 1572864 | 49316 | 2097152 |
| 30 | 15 (11) | 0.2 | 5896 | 8425520 | 185925 | 6389760 | 260645 | 16777216 |
| 30 | 15 (11) | 0.2 | 2096 | 2239982 | 27065 | 825344 | 42180 | 2097152 |
| 30 | 15 (12) | 0.2 | 3674 | 39928 | 6393 | 360448 | 9631 | 524288 |
| 30 | 15 (11) | 0.2 | 2388 | 552768 | 8623 | 425984 | 13641 | 524288 |
| 30 | 15 (11) | 0.2 | 1938 | 49388 | 11299 | 438272 | 27303 | 2097152 |
| 30 | 15 (13) | 0.2 | 4188 | 351374 | 18602 | 432776 | 38843 | 2097152 |
| 30 | 15 (12) | 0.2 | 2806 | 111632 | 3213 | 138240 | 5463 | 1048576 |
| 30 | 15 (12) | 0.2 | 3512 | 126464 | 16118 | 258048 | 11479 | 1048576 |
| 30 | 15 (10) | 0.2 | 1700 | 541498 | 5986 | 172032 | 8414 | 524288 |

Figure 17: Comparisons of random networks that exhibit CSI.

- *s* the number of splits and, in parentheses, the number of different variables on which the splits occur (different leaves can be split on the same variable).

- *p* the probability of splitting on a variable it is possible to split on.

- CBN size: the total size (summing over the size of the tables) of the contextual belief network representation.

- BN size: the total size of the factors for the corresponding belief network (i.e., assuming the probabilities are stored in tables).

- CVE time is the runtime (in msecs) of contextual variable elimination and CVE MTS is the maximum sum of the table sizes created for the elimination of a single variable.

- VE time: the runtime (in msecs) of variable elimination and VE MTS is the maximum table size created for the elimination of a single variable.

## References

Becker, A. and Geiger, D. (1996). A sufficiently fast algorithm for finding close to optimal junction trees, *in* E. Horvitz and F. Jensen (eds), *Proc. Twelfth Conf. on Uncertainty in Artificial Intelligence (UAI-96)*, Portland, OR, pp. 81–89.

Bertelè, U. and Brioschi, F. (1972). *Nonserial dynamic programming*, Vol. 91 of *Mathematics in Science and Engineering*, Academic Press.

Boutilier, C. (1997). Correlated action effects in decision theoretic regression, *in* Dan Geger and Prakash Shenoy (ed.), *Proceedings of the Thirteenth Annual Conference on Uncertainty in Artificial Intelligence (UAI–97)*, Providence, Rhode Island, pp. 30–37.

Boutilier, C., Dearden, R. and Goldszmidt, M. (1995). Exploiting structure in policy construction, *Proc. 14th International Joint Conf. on Artificial Intelligence (IJCAI-95)*, Montréal, Québec, pp. 1104–1111.

Boutilier, C., Friedman, N., Goldszmidt, M. and Koller, D. (1996). Context-specific independence in Bayesian networks, *in* E. Horvitz and F. Jensen (eds), *Proc. Twelfth Conf. on Uncertainty in Artificial Intelligence (UAI-96)*, Portland, OR, pp. 115–123.

Chickering, D. M., Heckerman, D. and Meek, C. (1997). A Bayesian approach to learning Bayesian networks with local structure, *Proc. Thirteenth Conf. on Uncertainty in Artificial Intelligence (UAI-97)*, pp. 80–89.

Dagum, P. and Luby, M. (1993). Approximating probabilistic inference in Bayesian belief networks is NP-hard, *Artificial Intelligence* **60**(1): 141–153.

D'Ambrosio (1995). Local expression languages for probabilistic dependence, *International Journal of Approximate Reasoning* **13**(1): 61–81.

| n | s | p | CBN Size | BN Size | CVE Time | CVE MTS | VE Time | VE MTS |
|---|---|---|---|---|---|---|---|---|
| 30 | 5 (3) | 0.2 | 1602 | 3658 | 662 | 49152 | 604 | 65536 |
| 30 | 5 (2) | 0.2 | 6108 | 7240 | 1583 | 131072 | 1945 | 131072 |
| 30 | 5 (2) | 0.2 | 17526 | 19632 | 6754 | 393216 | 8833 | 1048576 |
| 30 | 5 (2) | 0.2 | 892 | 4164 | 1604 | 81920 | 3119 | 131072 |
| 30 | 5 (1) | 0.2 | 1540 | 2640 | 261 | 16512 | 457 | 32768 |
| 30 | 5 (2) | 0.2 | 1156 | 5572 | 608 | 40960 | 816 | 65536 |
| 30 | 5 (2) | 0.2 | 2344 | 68676 | 12462 | 1048576 | 9370 | 1048576 |
| 30 | 5 (1) | 0.2 | 1406 | 7284 | 2197 | 81920 | 4021 | 262144 |
| 30 | 5 (3) | 0.2 | 7740 | 209652 | 17736 | 851968 | 18279 | 2097152 |
| 30 | 5 (2) | 0.2 | 4184 | 8838 | 8053 | 528384 | 16437 | 1048576 |
| 30 | 10 (3) | 0.2 | 3038 | 1119562 | 30669 | 1048576 | 53732 | 2097152 |
| 30 | 10 (3) | 0.2 | 888 | 5176 | 565 | 24576 | 2625 | 131072 |
| 30 | 10 (2) | 0.2 | 3372 | 4222408 | 21834 | 917504 | 108732 | 4194304 |
| 30 | 10 (3) | 0.2 | 2240 | 30968 | 20308 | 524800 | 129885 | 4194304 |
| 30 | 10 (2) | 0.2 | 1106 | 11660 | 741 | 32768 | 433 | 65536 |
| 30 | 10 (2) | 0.2 | 778 | 4582 | 274 | 18432 | 656 | 65536 |
| 30 | 10 (3) | 0.2 | 1888 | 72260 | 6659 | 262144 | 11986 | 524288 |
| 30 | 10 (2) | 0.2 | 9740 | 413892 | 11271 | 868352 | 179564 | 8388608 |
| 30 | 10 (2) | 0.2 | 1102 | 7744 | 313 | 16384 | 1395 | 65536 |
| 30 | 10 (3) | 0.2 | 4078 | 298438 | 61140 | 1048576 | 79858 | 4194304 |
| 30 | 15 (2) | 0.2 | 2710 | 50698 | 8845 | 524288 | 39265 | 2097152 |
| 30 | 15 (3) | 0.2 | 1246 | 84836 | 1935 | 90112 | 3652 | 262144 |
| 30 | 15 (3) | 0.2 | 2046 | 75956 | 54571 | 1310720 | 141386 | 8388608 |
| 30 | 15 (2) | 0.2 | 1588 | 138888 | 14280 | 458752 | 43059 | 2097152 |
| 30 | 15 (3) | 0.2 | 2260 | 20230 | 522 | 28672 | 1815 | 131072 |
| 30 | 15 (4) | 0.2 | 2842 | 67385366 | 322274 | 10485760 | 726989 | 33554432 |
| 30 | 15 (3) | 0.2 | 3074 | 533738 | 85480 | 2752512 | 89431 | 8388608 |
| 30 | 15 (3) | 0.2 | 1834 | 278426 | 18560 | 753664 | 43554 | 1048576 |
| 30 | 15 (3) | 0.2 | 4362 | 209186 | 22872 | 1441792 | 131704 | 4194304 |
| 30 | 15 (2) | 0.2 | 3142 | 151160 | 4476 | 164096 | 26426 | 1048576 |

Figure 18: Some of the details of data from Figure 12 (biased towards fewer different variables)

Darwiche, A. (1995). Conditioning algorithms for exact and approximate inference in causal networks, *in* P. Besnard and S. Hanks (ed.), *Proc. Eleventh Conf. on Uncertainty in Artificial Intelligence (UAI-95)*, Montreal, Quebec, pp. 99–107.

Dearden, R. and Boutilier, C. (1997). Abstraction and approximate decision theoretic planning, *Artificial Intelligence* **89**(1): 219–283.

Dechter, R. (1996). Bucket elimination: A unifying framework for probabilistic inference, *in* E. Horvitz and F. Jensen (eds), *Proc. Twelfth Conf. on Uncertainty in Artificial Intelligence (UAI-96)*, Portland, OR, pp. 211–219.

Díez, F. (1996). Local conditioning in Bayesian networks, *Artificial Intelligence* **87**(1–2): 1–20.

Friedman, N. and Goldszmidt, M. (1996). Learning Bayesian networks with local structure, *Proc. Twelfth Conf. on Uncertainty in Artificial Intelligence (UAI-96)*, pp. 252–262.

Geiger, D. and Heckerman, D. (1996). Knowledge representation and inference in similarity networks and Bayesian multinets, *Artificial Intelligence* **82**: 45–74.

Guestrin, C., Venkataraman, S. and Koller, D. (2002). Context specific multiagent coordination and planning with factored MDPs, *The Eighteenth National Conference on Artificial Intelligence (AAAI-2002)*, Edmonton, Canada.

Heckerman, D. and Breese, J. (1994). A new look at causal independence, *Proc. of the Tenth Conference on Uncertainty in Artificial Intelligence*, pp. 286–292.

Jensen, F. V., Kjærulff, U., Olesen, K. G. and Pedersen, J. (1989). Et forprojekt til et ekspertsystem for drift af spildevandsrensning (an expert system for control of waste water treatment — a pilot project), *Technical report*, Judex Datasystemer A/S, Aalborg, Denmark. In Danish.

Jensen, F. V., Lauritzen, S. L. and Olesen, K. G. (1990). Bayesian updating in causal probabilistic networks by local computations, *Computational Statistics Quarterly* **4**: 269–282.

Kjærulff, U. (1990). Triangulation of graphs - algorithms giving small total state space, *Technical Report R 90-09*, Department of Mathematics and Computer Science, Strandvejen, DK 9000 Aalborg, Denmark.

Lauritzen, S. L. and Spiegelhalter, D. J. (1988). Local computations with probabilities on graphical structures and their application to expert systems, *Journal of the Royal Statistical Society, Series B* **50**(2): 157–224.

Neal, R. (1992). Connectionist learning of belief networks, *Artificial Intelligence* **56**: 71–113.

Pearl, J. (1988). *Probabilistic Reasoning in Intelligent Systems: Networks of Plausible Inference*, Morgan Kaufmann, San Mateo, CA.

Poole, D. (1993). Probabilistic Horn abduction and Bayesian networks, *Artificial Intelligence* **64**(1): 81–129.

Poole, D. (1995). Exploiting the rule structure for decision making within the independent choice logic, *in* P. Besnard and S. Hanks (eds), *Proc. Eleventh Conf. on Uncertainty in Artificial Intelligence (UAI-95)*, Montréal, Québec, pp. 454–463.

Poole, D. (1997). Probabilistic partial evaluation: Exploiting rule structure in probabilistic inference, *Proc. 15th International Joint Conf. on Artificial Intelligence (IJCAI-97)*, Nagoya, Japan, pp. 1284–1291.

Poole, D. (1998). Context-specific approximation in probabilistic inference, *in* G.F. Cooper and S. Moral (ed.), *Proc. Fourteenth Conf. on Uncertainty in Artificial Intelligence*, Madison, WI, pp. 447–454.

Saul, L., Jaakkola, T. and Jordan, M. (1996). Mean field theory for sigmoid belief networks, *Journal of Artificial Intelligence Research* **4**: 61–76.

Shachter, R. D., D'Ambrosio, B. D. and Del Favero, B. D. (1990). Symbolic probabilistic inference in belief networks, *Proc. 8th National Conference on Artificial Intelligence*, MIT Press, Boston, pp. 126–131.

Smith, J. E., Holtzman, S. and Matheson, J. E. (1993). Structuring conditional relationships in influence diagrams, *Operations Research* **41**(2): 280–297.

Tung, L. (2002). *A clique tree algorithm exploiting context-specific independence*, Master's thesis, Department of Computer Science, University of British Columbia.

Zhang, N. L. (1998). Inference in Bayesian networks: The role of context-specific independence, *Technical Report HKUST-CS98-09*, Department of Computer Science, Hong Kong University of Science and Technology.

Zhang, N. L. and Poole, D. (1999). On the role of context-specific independence in probabilistic reasoning, *Proc. 16th International Joint Conf. on Artificial Intelligence (IJCAI-99)*, Stockholm, Sweden, pp. 1288–1293.

Zhang, N. and Poole, D. (1994). A simple approach to Bayesian network computations, *Proc. of the Tenth Canadian Conference on Artificial Intelligence*, pp. 171–178.

Zhang, N. and Poole, D. (1996). Exploiting causal independence in Bayesian network inference, *Journal of Artificial Intelligence Research* **5**: 301–328.